%% file: ms.tex
\documentclass[letterpaper]{article}
\usepackage{aaai23}
\usepackage{times}
\usepackage{helvet}
\usepackage{courier}
\usepackage[hyphens]{url}
\usepackage{graphicx}
\usepackage{natbib}
\usepackage{caption}
\usepackage{algorithm}
\usepackage[noend]{algpseudocode}

\input{math_commands.tex}

\usepackage{booktabs}       
\usepackage{amsmath,amsfonts,bm}       
\usepackage{nicefrac}       
\usepackage{microtype}      
\usepackage{xcolor}         
\usepackage{subfigure}
\usepackage[para]{threeparttable} 
\usepackage{multirow}
\usepackage{array} 
\usepackage{mathtools}

\usepackage{hyperref}
\colorlet{mydarkblue}{blue!60!black}
\hypersetup{
    pdftitle={Learning Similarity Metrics for Volumetric Simulations with Multiscale CNNs},
    pdfauthor={Georg Kohl, Li-Wei Chen, Nils Thuerey},
    pdfsubject={Submission to ArXiv},
    pdfkeywords={metric learning, CNNs, PDEs, numerical simulation, perceptual evaluation, physics simulation},
    colorlinks=true,
    linkcolor=mydarkblue,
    citecolor=mydarkblue,
    filecolor=mydarkblue,
    urlcolor=mydarkblue,
}
\usepackage{doi}

\makeatletter
\newcommand\smallfontsize{\@setfontsize\smallfontsize{9.0}{10.0}}
\makeatother

\newcommand{\best}[1]{\textbf{\textcolor{green!25!black}{\underline{#1}}}}
\newcommand{\bestErr}[1]{\textbf{\textcolor{green!25!black}{#1}}}
\newcommand{\different}[1]{\textcolor{white!40!black}{#1}}

\title{Learning Similarity Metrics for Volumetric Simulations with Multiscale CNNs}
\author {
    Georg Kohl, 
    Li-Wei Chen, 
    Nils Thuerey
}
\affiliations {
    Technical University of Munich\\
    \{georg.kohl, liwei.chen, nils.thuerey\}@tum.de
}

\begin{document}

\maketitle

\begin{abstract}
Simulations that produce three-dimensional data are ubiquitous in science, ranging from fluid flows to plasma physics. We propose a similarity model based on entropy, which allows for the creation of physically meaningful ground truth distances for the similarity assessment of scalar and vectorial data, produced from transport and motion-based simulations. Utilizing two data acquisition methods derived from this model, we create collections of fields from numerical PDE solvers and existing simulation data repositories. Furthermore, a multiscale CNN architecture that computes a volumetric similarity metric (\textit{VolSiM}) is proposed. To the best of our knowledge this is the first learning method inherently designed to address the challenges arising for the similarity assessment of high-dimensional simulation data. Additionally, the tradeoff between a large batch size and an accurate correlation computation for correlation-based loss functions is investigated, and the metric's invariance with respect to rotation and scale operations is analyzed. Finally, the robustness and generalization of \textit{VolSiM} is evaluated on a large range of test data, as well as a particularly challenging turbulence case study, that is close to potential real-world applications.
\end{abstract}

\section{Introduction}
Making comparisons is a fundamental operation that is essential for any kind of computation. This is especially true for the simulation of physical phenomena, as we are often interested in comparing simulations against other types of models or measurements from a physical system. Mathematically, such comparisons require metric functions that determine scalar distance values as a similarity assessment. A fundamental problem is that traditional comparisons are typically based on simple, element-wise metrics like the $L^1$ or $L^2$ distances, due to their computational simplicity and a lack of alternatives. Such metrics can work reliably for systems with few elements of interest, e.g.~if we want to analyze the position of a moving object at different points in time, matching our intuitive understanding of distances. However, more complex physical problems often exhibit large numbers of degrees of freedom, and strong dependencies between elements in their solutions. Those coherences should be considered when comparing physical data, but element-wise operations by definition ignore such interactions between elements.
With the curse of dimensionality, this situation becomes significantly worse for systems that are modeled with dense grid data, as the number of interactions grows exponentially with a linearly increasing number elements. Such data representations are widely used, e.g. for medical blood flow simulations \cite{olufsen2000_Numerical}, climate and weather predictions \cite{stocker2014_Climate}, and even the famous unsolved problem of turbulence \cite{holmes2012_Turbulence}. Another downside of element-wise metrics is that each element is weighted equally, which is typically suboptimal; e.g.~smoke plumes behave differently along the vertical dimension due to gravity or buoyancy, and small key features like vortices are more indicative of the fluid's general behavior than large areas of near constant flow \cite{pope2000_Turbulent}.

In the image domain, neural networks have been employed for similarity models that can consider larger structures, typically via training with class labels that provide semantics, or with data that encodes human perception. Similarly, physical systems exhibit spatial and temporal coherence due to the underlying laws of physics that can be utilized. In contrast to previous work on simulation data \cite{kohl2020_Learning}, we derive an entropy-based similarity model to robustly learn similarity assessments of scalar and vectorial volumetric data. Overall, our work contributes the following:
\begin{itemize}   
    \item We propose a novel similarity model based on the entropy of physical systems. It is employed to synthesize sequences of volumetric physical fields suitable for metric learning.
    \item We show that our Siamese multiscale feature network results in a stable metric that successfully generalizes to new physical phenomena. To the best of our knowledge this is the first learned metric inherently designed for the similarity assessment of volumetric fields.
    \item The metric is employed to analyze turbulence in a case study, and its invariance to rotation and scale are evaluated. In addition, we analyze correlation-based loss functions with respect to their tradeoff between batch size and accuracy of correlation computation.
\end{itemize}
The central application of the proposed \textit{VolSiM} metric is the similarity assessment of new physical simulation methods, numerical or learning-based, against a known ground truth.\footnote{Our source code, datasets, and ready-to-use models are available at \url{https://github.com/tum-pbs/VOLSIM}.} This ground truth can take the form of measurements, higher resolution simulations, or existing models. Similar to perceptual losses for computer vision tasks, the trained metric can also be used as a differentiable similarity loss for various physical problems. We refer to \citet{thuerey2021_Physicsbased} for an overview of such problems and different learning methods to approach them.

\section{Related Work}
Apart from simple $L^n$ distances, the two metrics peak signal-to-noise ratio (PSNR) and structural similarity index (SSIM) from \citeauthor{wang2004_Image} are commonly used across disciplines for the similarity assessment of data. Similar to the underlying $L^2$ distance, PSNR shares the issues of element-wise metrics \cite{huynh-thu2008_Scope,huynh-thu2012_Accuracy}.
SSIM computes a more intricate function, but it was shown to be closely related to PSNR \cite{hore2010_Image} and thus has similar problems \cite{nilsson2020_Understanding}. Furthermore, statistical tools like the Pearson correlation coefficient PCC \cite{pearson1920_Notes} and Spearman's rank correlation coefficient SRCC \cite{spearman1904_Proof} can be employed as a simple similarity measurement.
There are several learning-based metrics specialized for different domains such as rendered \cite{andersson2020_FLIP} and natural images \cite{bosse2016_Neural}, interior object design \cite{bell2015_Learning}, audio \cite{avgoustinakis2020_Audiobased}, and haptic signals \cite{kumari2019_PerceptNet}.

Especially for images, similarity measurements have been approached in various ways, but mostly by combining deep embeddings as perceptually more accurate metrics \cite{prashnani2018_PieAPP,talebi2018_Learned}. These metrics can be employed for various applications such as image super-resolution \cite{johnson2016_Perceptual} or generative tasks \cite{dosovitskiy2016_Generating}. Traditional metric learning for images typically works in one of two ways: Either, the training is directly supervised by learning from manually created labels, e.g.~via two-alternative forced choice where humans pick the most similar option to a reference \cite{zhang2018_Unreasonable}, or the training is indirectly semi-supervised through images with class labels and a contrastive loss \cite{chopra2005_Learning, hadsell2006_Dimensionality}. In that case, triplets of reference, same class image, and other class images are sampled, and the corresponding latent space representations are pulled together or pushed apart. We refer to \citet{roth2020_Revisiting} for an overview of different training strategies for learned image metrics. 
In addition, we study the behavior of invariance and equivariance to different transformations, which was targeted previously for rotational symmetries \cite{weiler2018_3D, chidester2019_Rotation} and improved generalization \cite{wang2021_Incorporating}.

Similarity metrics for simulation data have not been studied extensively yet. Siamese networks for finding similar fluid descriptors have been applied to smoke flow synthesis, where a highly accurate similarity assessment is not necessary \cite{chu2017_DataDriven}. Um et al.~(\citeyear{um2017_Perceptual,um2021_Spot}) used crowd-sourced user studies for the similarity assessment of liquid simulations which rely on relatively slow and expensive human evaluations. Scalar 2D simulation data was previously compared with a learned metric using a Siamese network \cite{kohl2020_Learning}, but we overcome methodical weaknesses and improve upon the performance of their work. Their \textit{LSiM} method relies on a basic feature extractor based on common classification CNNs, does not account for the long-term behavior of different systems with respect to entropy via a similarity model during training, and employs a simple heuristic to generate suitable data sequences.

\section{Modeling Similarity of Simulations} \label{sec: simulation similarity}
To formulate our methodology for learning similarity metrics that target dissipative physical systems, we turn to the fundamental quantity of entropy. The second law of thermodynamics states that the entropy $S$ of a closed physical system never decreases, thus $\Delta S \geq 0$. In the following, we make the reasonable assumption that the behavior of the system is continuous and non-oscillating, and that $\Delta S > 0$.\footnote{These assumptions are required to create sequences with meaningful ground truth distances below in Sec.~\ref{sec: sequence creation}.}
Eq.~\ref{eq: entropy 1} is the Boltzmann equation from statistical mechanics \cite{boltzmann1866_Uber}, that describes $S$ in terms of the Boltzmann constant $k_b$ and the number of microstates $W$ of a system.\footnote{We do not have any a priori information about the distribution of the likelihood of each microstate in a general physical system. Thus, the Boltzmann entropy definition which assumes a uniform microstate distribution is used in the following, instead of more generic entropy models such as the Gibbs or Shannon entropy.} 
\begin{equation}
    S = k_B \log(W) \label{eq: entropy 1}
\end{equation}
Since entropy only depends on a single system state, it can be reformulated to take the relative change between two states into account. From an information-theoretical perspective, this is related to using Shannon entropy \cite{shannon1948_Mathematical} as a diversity measure, as done by \citet{renyi1961_Measures}.
Given a sequence of states $s_0, s_1, \dotsc, s_n$, we define the relative entropy 
\begin{equation}
    \tilde{S}(\vs) = k \log(10^c \vw_s). \label{eq: entropy 2}
\end{equation}
Here, $\vw_s$ is the monotonically increasing, relative number of microstates defined as $0$ for $s_0$ and as $1$ for $s_n$. $10^c > 0$ is a system-dependent factor that determines how quickly the number of microstates increases, i.e.~it represents the speed at which different processes decorrelate. As the properties of similarity metrics dictate that distances are always non-negative and only zero for identical states, the lower bound in  Eq.~\ref{eq: entropy 2} is adjusted to $0$, leading to a first similarity model $\hat{D}(\vs) = k \log(10^c\,\vw_s + 1).$ Finally, relative similarities are equivalent up to a multiplicative constant, and thus we can freely choose $k$. Choosing $k = 1 / (\log{10^c + 1})$ leads to the full similarity model
\begin{equation}
    \operatorname{D}(\vs) = \frac{\log(10^c\,\vw_s + 1)}{\log(10^c + 1)}. \label{eq: entropy 3}
\end{equation}
For a sequence $\vs$, it predicts the overall similarity behavior between the reference $s_0$ and the other states with respect to entropy, given the relative number of microstates $\vw_s$ and the system decorrelation speed $c$.
\begin{figure}[ht]
    \centering
    \includegraphics[width=0.47\textwidth]{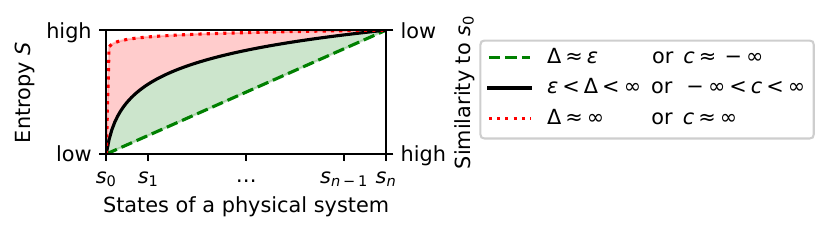}
    \caption{Idealized model of the behavior of entropy and similarity for different $\Delta$ or different $c$, respectively.}
    \label{fig: entropy behavior}
\end{figure}

Fig.~\ref{fig: entropy behavior} illustrates the connection between the logarithmically increasing entropy and the proposed similarity model for a state sequence with fixed length $n$. Here, $\Delta$ denotes the magnitude of change between the individual sequence states which is directly related to $\vw_s$, and $c$ is the decorrelation speed of the system that produced the sequence. In the following, we will refer to the property of a sequence being informative with respect to a pairwise similarity analysis as \emph{difficulty}. Sequences that align with the red dotted curve contain little information as they are dissimilar to $s_0$ too quickly, either because the original system decorrelates too fast or because the difference between each state is too large (\emph{high difficulty}). On the other hand, sequences like the green dashed curve are also not ideal as they change too little, and a larger non-practical sequence length would be necessary to cover long-term effects (\emph{low difficulty}). Ideally, a sequence $\vs$ employed for learning tasks should evenly exhibit both regimes as well as intermediate ones, as indicated by the black curve. The central challenges now become finding sequences with a suitable magnitude of $\Delta$, determining $c$, and assigning distances $\vd$ to pairs from the sequence.

\section{Sequence Creation} \label{sec: sequence creation}
To create a sequence $s_0, s_1, \dotsc, s_n$ within a controlled environment, we make use of the knowledge about the underlying physical processes: We either employ direct changes, based on spatial or temporal coherences to $s_0$, or use changes to the initial conditions of the process that lead to $s_0$. As we can neither directly determine $c$ nor $\vd$ at this point, we propose to use proxies for them during the sequence generation. Initially, this allows for finding sequences that roughly fall in a suitable difficulty range, and accurate values can be computed afterwards. Here, we use the mean squared error (MSE) as a proxy distance function and the PCC to determine $c$, to iteratively update $\Delta$ to a suitable range.

\begin{figure}[!b]
    \centering
    \includegraphics[width=0.46\textwidth]{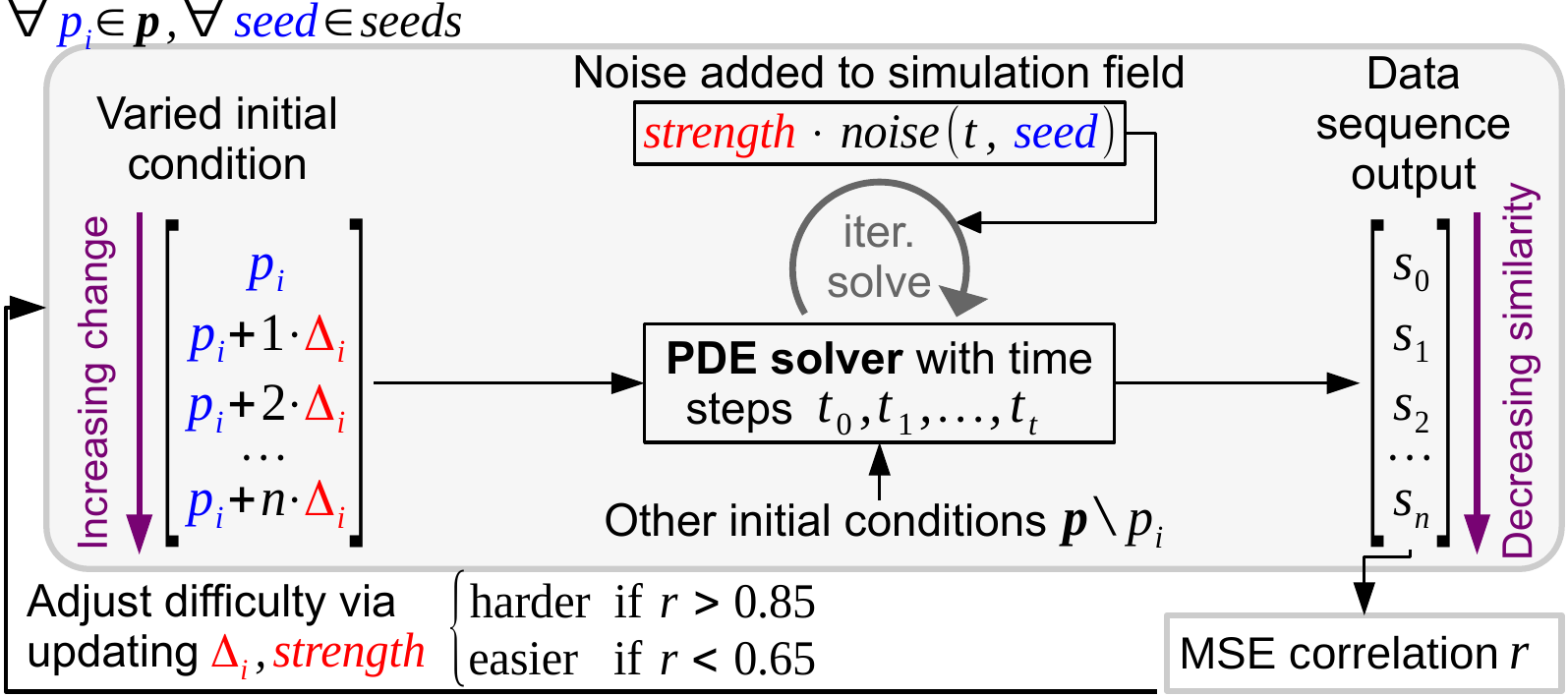}

    \vspace{0.15cm}

    \includegraphics[width=0.46\textwidth]{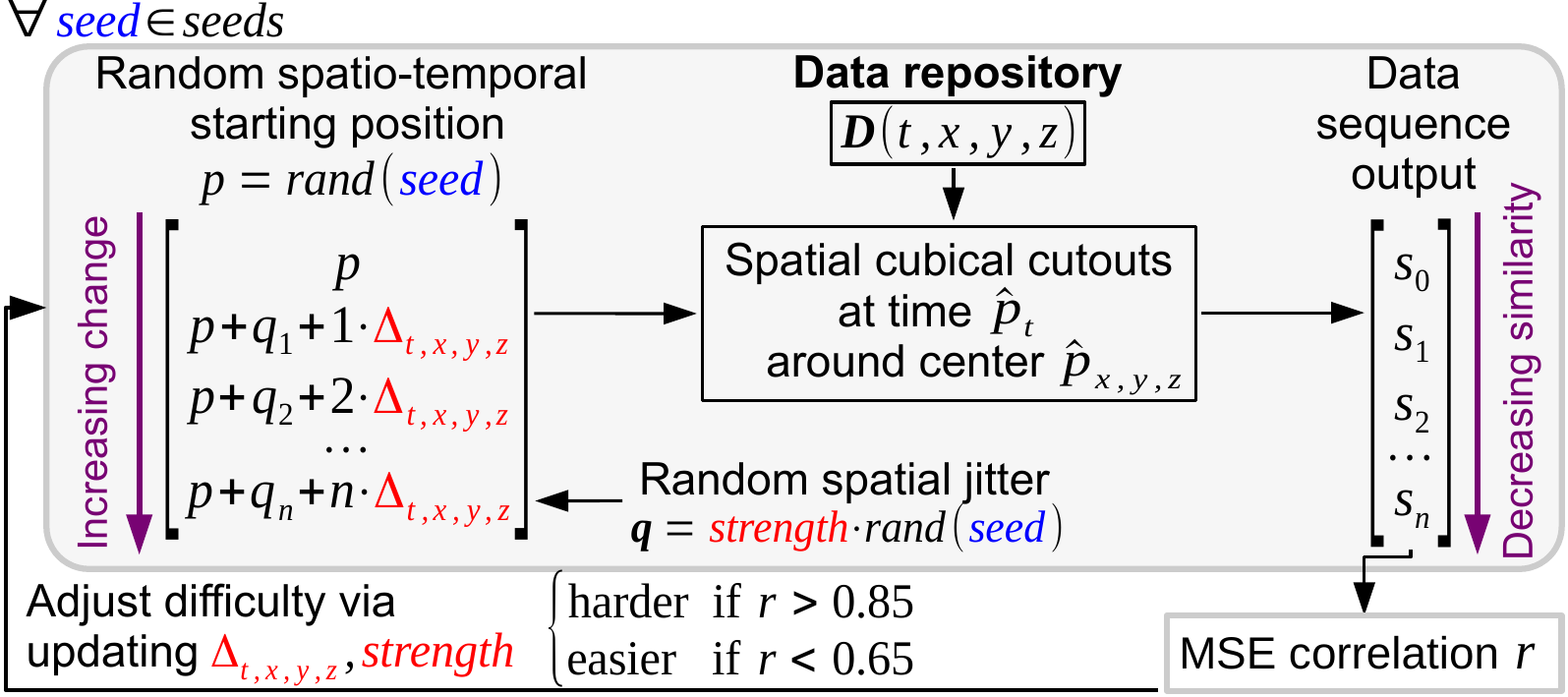}

    \caption{Iteration schemes to calibrate and create data sequences of decreasing similarity. Variation from the reference state can be introduced via the initial conditions of a numerical PDE simulation (method [A], top), or via spatio-temporal data changes on data from a repository (method [B], bottom).}
    \label{fig: data iteration schemes}
\end{figure}

Given any value of $\Delta$ and a corresponding sequence, pairwise proxy distances\footnote{To keep the notation clear and concise, sequentially indexing the distance vectors $\vd^{\Delta}$ and $\vw_s$ with i and j is omitted here.} between the sequence elements are computed $\vd^{\Delta} = \operatorname{MSE}(s_i,s_j)$ and min-max normalized to $[0,1]$. Next, we determine a distance sequence corresponding to the physical changes over the states, which we model as a simple linear increase over the sequence $\vw_s = (j-i)/n$ following \cite{kohl2020_Learning}. To indirectly determine $c$, we compare how both distance sequences differ in terms of the PCC as $r=\operatorname{PCC}(\vd^{\Delta}, \vw_s)$. We empirically determined that correlations between $0.65$ and $0.85$ work well for all cases we considered. In practice, the network stops learning effectively for lower correlation values as states are too different, while sequences with higher values reduce generalization as a simple metric is sufficient to describe them. Using these thresholds, we propose two semi-automatic iterative methods to create data, depending on the method to introduce variations to a given state (see Fig.~\ref{fig: data iteration schemes}). Both methods sample a small set of sequences to calibrate $\Delta$ to a suitable magnitude and use that value for the full data set. Compared to strictly sampling every sequence, this method is computationally significantly more efficient as less sampling is needed, and it results in a more natural data distribution.

\paragraph{[A] Variations from Initial Conditions of Simulations}
Given a numerical PDE solver and a set of initial conditions or parameters $\vp$, the solver computes a solution to the PDE over the time steps $t_0, t_1, \dotsc, t_t$. To create a larger number of different sequences, we make the systems non-deterministic by adding noise to a simulation field and randomly generating the initial conditions from a given range. Adjusting \emph{one} of the parameters $p_i$ in steps with a small perturbation $\Delta_i$, allows for the creation of a sequence $s_0, s_1, \dotsc, s_n$ with decreasing similarity to the unperturbed simulation output $s_0$. This is repeated for every suitable parameter in $\vp$, and the corresponding $\Delta$ is updated individually until the targeted MSE correlation range is reached. The global noise strength factor also influences the difficulty and can be updated.
\paragraph{[B] Variations from Spatio-temporal Coherences}
For a source $\tD$ of volumetric spatio-temporal data without access to a solver, we rely on a larger spatial and/or temporal dimension than the one required for a sequence. We start at a random spatio-temporal position $p$ to extract a cubical spatial area $s_0$ around it. $p$ can be repeatedly translated in space and/or time by $\Delta_{t,x,y,z}$ to create a sequence $s_0, s_1, \dotsc, s_n$ of decreasing similarity. Note that individual translations in space or time should be preferred if possible. Using different starts leads to new sequences, as long as enough diverse data is available. It is possible to add some global random perturbations $q$ to the positions to further increase the difficulty.

\paragraph{Data Sets}
To create training data with method [A], we utilize solvers for a basic Advection-Diffusion model (\texttt{Adv}), Burgers' equation (\texttt{Bur}) with an additional viscosity term, and the full Navier-Stokes equations via a Eulerian smoke simulation (\texttt{Smo}) and a hybrid Eulerian-Lagrangian liquid simulation (\texttt{Liq}). The corresponding validation sets are generated with a separate set of random seeds. Furthermore, we use adjusted versions of the noise integration for two test sets, by adding noise to the density instead of the velocity in the Advection-Diffusion model (\texttt{AdvD}) and overlaying background noise in the liquid simulation (\texttt{LiqN}).

We create seven test sets via method [B]. Four come from the Johns Hopkins Turbulence Database JHTDB \cite{perlman2007_Data} that contains a large amount of direct numerical simulation (DNS) data, where each is based on a subset of the JHTDB and features different characteristics: iso\-tro\-pic turbulence (\texttt{Iso}), a channel flow (\texttt{Cha}), mag\-ne\-to-hy\-dro\-dy\-na\-mic turbulence (\texttt{Mhd}), and a tran\-si\-tio\-nal boundary layer (\texttt{Tra}). Since turbulence contains structures of interest across all length scales, we additionally randomly stride or interpolate the query points for scale factors in $[0.25, 4]$ to create sequences of different physical size. One additional test set (\texttt{SF}) via temporal translations is based on ScalarFlow \cite{eckert2019_ScalarFlow}, consisting of 3D reconstructions of real smoke plumes. Furthermore, method [B] is slightly modified for two synthetic test sets: Instead of using a data repository, we procedurally synthesize spatial fields: We employ linearly moving randomized shapes (\texttt{Sha}), and randomized damped waves (\texttt{Wav}) of the general form $f(x)=cos(x)*e^{-x}$. All data was gathered in sequences with $n=10$ at resolution $128^3$, and downsampled to $64^3$ for computational efficiency during training and evaluations.

\paragraph{Determining $c$}
For each calibrated sequence, we can now more accurately estimate $c$. As $c$ corresponds to the decorrelation speed of the system, we choose Pearson's distance $d^{\Delta}_i = 1 - \left|\operatorname{PCC}(s_0,s_i)\right|$ as a distance proxy here. $c$ is determine via standard unbounded least-squares optimization from the similarity model in Eq.~\ref{eq: entropy 3} as $c = \operatorname{arg\,min}_c \frac{\log(10^c\,\vd^\Delta + 1)}{\log(10^c + 1)}$.

\section{Learning a Distance Function} \label{sec: distance function}
Given the calibrated sequences $\vs$ of different physical systems with elements $s_0, s_1, \dotsc, s_n$, the corresponding value of $c$, and the pairwise physical target distance sequence $\vw_s = (j-i)/n$, we can now formulate a semi-supervised learning problem: We train a neural network $m$ that receives pairs from $\vs$ as an input, and outputs scalar distances $\vd$ for each pair. These predictions are trained against ground truth distances $\vg = \frac{\log(10^c\,\vw_s + 1)}{\log(10^c + 1)}$. Note that $\vg$ originates from the sequence order determined by our data generation approach, transformed with a non-linear transformation according to the entropy-based similarity model. This technique incorporates the underlying physical behavior by accounting for the decorrelation speed over the sequence, compared to adding variations in a post-process (as commonly done in the domain of images, e.g.~by \citet{ponomarenko2015_Image}). To train the metric network, the correlation loss function in Eq.~\ref{eq: correlation loss} below compares $\vd$ to $\vg$ and provides gradients.

\paragraph{Network Structure}
For our method, we generally follow the established Siamese network structure, that was originally proposed for 2D domains \cite{zhang2018_Unreasonable}: First, two inputs are embedded in a latent space using a CNN as a feature extractor. The Siamese structure means that the weights are shared, which ensures the mathematical requirements for a pseudo-metric \cite{kohl2020_Learning}. Next, the features from all layers are normalized and compared with an element-wise comparison like an absolute or squared difference. Finally, this difference is aggregated with sum, mean, and learned weighted average functions. To compute the proposed \textit{VolSiM} metric that compares inherently more complex 3D data, changes to this framework are proposed below.

\begin{figure}[!b]
    \centering
    \includegraphics[width=0.47\textwidth]{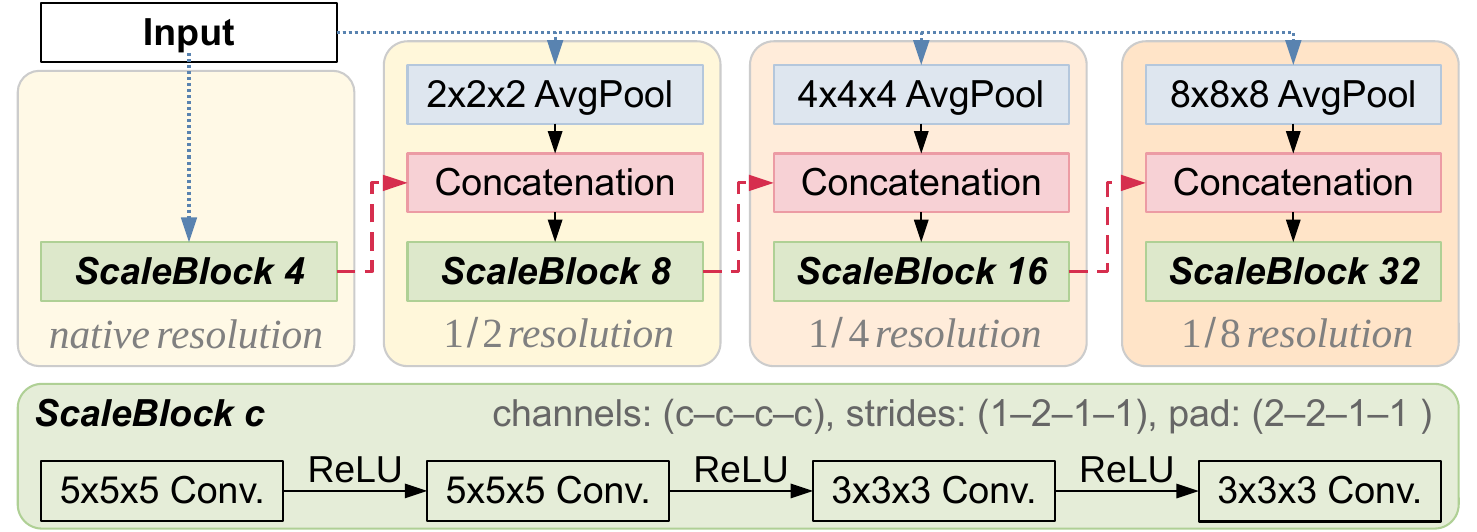}
    \caption{Standard Conv+ReLU blocks (bottom) are interwoven with input and resolution connections (blue dotted and red dashed), to form the combined network architecture (top) with about $350k$ weights.}
    \label{fig: multiscale network}
\end{figure}

\paragraph{Multiscale Network}
Scale is important for a reliable similarity assessment, since physical systems often exhibit self-similar behavior that does not significantly change across scales, as indicated by the large number of dimensionless quantities in physics. Generally, scaling a data pair should not alter its similarity, and networks can learn such an invariance to scale most effectively by processing data at different scales. One example where this is crucial is the energy cascade in turbulence \cite{pope2000_Turbulent}, which is also analyzed in our case study below. For learned image metrics, this invariance is also useful (but less crucial), and often introduced with large strides and kernels in the convolutions, e.g.~via a feature extractor based on AlexNet \cite{zhang2018_Unreasonable}. In fact, our experiments with similar architectures showed, that models with large strides and kernels generally perform better than models that modify the scale over the course of the network to a lesser extent. However, we propose to directly encode this scale structure in a multiscale architecture for a more accurate similarity assessment, and a network with a smaller resource footprint.

Fig.~\ref{fig: multiscale network} shows the proposed fully convolutional network: Four scale blocks individually process the input on increasingly smaller scales, where each block follows the same layer structure, but deeper blocks effectively cover a significantly larger volume due to the reduced input resolutions. Deeper architectures can model complex functions more easily, so we additionally include resolution connections from each scale block to the next resolution level via concatenation. Effectively, the network learns a mixture of connected deep features and similar representations across scales as a result.

\paragraph{Training and Evaluation}
To increase the model's robustness during training, we used the following data augmentations for each sequence: the data is normalized to $[-1,1]$, and together randomly flipped and rotated in increments of 90\textdegree~around a random axis. The velocity channels are randomly swapped to prevent directional biases from some simulations, while scalar data is extended to the three input channels via repetition. For inference, only the normalization operation and the repetition of scalar data is performed. The final metric model was trained with the Adam optimizer with a learning rate of $10^{-4}$ for 30 epochs via early stopping. To determine the accuracy of any metric during inference in the following, we compute the SRCC between the distance predictions of the metric $\vd$ and the ground truth $\vw_s$, where a value closer to $1$ indicates a better reconstruction.\footnote{This is equivalent to $\operatorname{SRCC}(\vd, \vg)$, since the SRCC measures monotonic relationships and is not affect by monotonic transformations, but using $\vw_s$ is more efficient and has numerical benefits.}

\paragraph{Loss Function}
Given predicted distances $\vd$ and a ground truth $\vg$ of size $n$, we train our metric networks with the loss
\begin{equation}
\begin{aligned}
    &\operatorname{L}(\vd, \vg) = \lambda_1 (\vd - \vg)^2 + \lambda_2 (1 - r) \\
    &\text{where} \;
    r = \frac{\sum_{i=1}^n (d_i - \bar{d}) \, (g_i - \bar{g})}
    { \sqrt {\sum_{i=1}^n (d_i - \bar{d})^2 } \, \sqrt{\vphantom{\bar{d}} \sum_{i=1}^n (g_i - \bar{g})^2 } }.
    \label{eq: correlation loss}
\end{aligned}
\end{equation}
consisting of a weighted combination of an MSE and an inverted correlation term $r$, where $\bar{d}$ and $\bar{g}$ denote the mean. While the formulation follows existing work \cite{kohl2020_Learning}, it is important to note that $\vg$ is computed by our similarity model from Sec.~\ref{sec: simulation similarity}, and below we introduce a slicing technique to apply this loss formulation to high-dimensional data sets.

To successfully train a neural network, Eq.~\ref{eq: correlation loss} requires a trade-off: A large batch size $b$ is useful to improve training stability via less random gradients for optimization. Similarly, a sufficiently large value of $n$ is required to keep the correlation values accurate and stable. However, with finite amounts of memory, choosing large values for $n$ and $b$ is not possible in practice. Especially so for 3D cases, where a single sample can already be memory intensive. In general, $n$ is implicitly determined by the length of the created sequences via the number of possible pairs. Thus, we provide an analysis how the correlation can be approximated in multiple steps for a fixed $n$, to allow for increasing $b$ in return. In the following, the batch dimension is not explicitly shown, but all expressions can be extended with a vectorized first dimension. The full distance vectors $\vd$ and $\vg$ are split in slices with $v$ elements, where $v$ should be a proper divisor of $n$. For any slice $k$, we can compute a partial correlation $r_k$ with
\begin{equation}
     r_k = \frac{\sum_{i=k}^{k+v} (d_i - \bar{d}) \, (g_i - \bar{g})}
    { \sqrt {\vphantom{\bar{g}} \sum_{i=k}^{k+v} (d_i - \bar{d})^2 } \, \sqrt{ \sum_{i=k}^{k+v} (g_i - \bar{g})^2 } }.
    \label{eq: correlation slicing}
\end{equation}
Note that this is only an approximation, and choosing larger values of $v$ for a given $b$ is always beneficial, if sufficient memory is available. For all slices, the gradients are accumulated during backpropagation since other aggregations would required a computational graph of the original, impractical size. Eq.~\ref{eq: correlation slicing} still requires the computation of the means $\bar{d}$ and $\bar{g}$ as a pre-process over all samples. Both can be approximated with the running means $\tilde{d}$ and $\tilde{g}$ for efficiency ($RM$). For small values of $v$, the slicing results in very coarse, unstable correlation results. To alleviate that, it is possible to use a running mean over all previous values $\tilde{r}_k = (1/k) ( r_k + \sum_{l=1}^{k-1} r_l )$. This aggregation ($AG$) can stabilize the gradients of individual $r_k$ as they converge to the true correlation value.

\begin{figure}[!tb]
    \centering
    \includegraphics[width=0.47\textwidth]{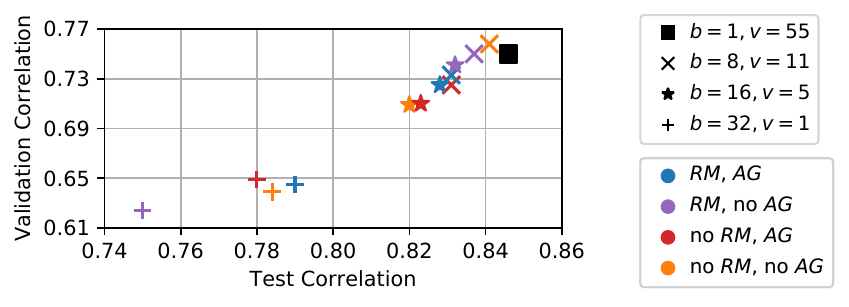}
    \caption{Combined validation and test performance for different batch sizes $b$ and slicing values $v$ (markers), and the usage of running sample mean $RM$ and correlation aggregation $AG$ (colors).}
    \label{fig: loss comparison}
\end{figure}

Fig.~\ref{fig: loss comparison} displays the resulting performance on our data, when training with different combinations of $b$, $v$, $RM$, and $AG$. All models exhibit similar levels of memory consumption and were trained with the same random training seed. When comparing models with and without $RM$ both are on par in most cases, even though computation times for a running mean are about $20\%$ lower. Networks with and without $AG$ generalize similarly, however, models with the aggregation exhibit less fluctuations during optimization, leading to an easier training process. Overall, this experiment demonstrates that choosing larger $v$ consistently leads to better results (marker shape), so more accurate correlations are beneficial over a large batch size $b$ in memory-limited scenarios. Thus, we use $b=1$ and $v=55$ for the final model.

\begin{table*}[ht]
    \smallfontsize
    \centering
    
    \begin{threeparttable}
    \begin{tabular}[b]{l c c c c || c c c c c c c c c | c}
        \toprule
        \multirow{2}{*}[-1.3mm]{} & \multicolumn{4}{c ||}{\bf Validation data sets} & \multicolumn{10}{c}{\bf Test data sets} \\
        \cmidrule(lr){2-5} \cmidrule(lr){6-15}
        & \multicolumn{4}{c }{Simulated} & \multicolumn{2}{c |}{Simulated} & \multicolumn{2}{c |}{Generated} & \multicolumn{4}{c |}{JHTDB\tnote{a}} & SF\tnote{b} & \tnote{c} \\
        \cmidrule(lr){2-5} \cmidrule(lr){6-15}
        \bf Metric & \texttt{Adv} & \texttt{Bur} & \texttt{Liq} & \texttt{Smo} & \texttt{AdvD} & \texttt{LiqN} & \texttt{Sha} & \texttt{Wav} & \texttt{Iso} & \texttt{Cha} & \texttt{Mhd} & \texttt{Tra} & \texttt{SF} & \texttt{All}\\
        \cmidrule(lr){1-15}

        \it MSE           & 0.61 & 0.70 & 0.51 & 0.68  & 0.77 & 0.76 & 0.75 & 0.65  & 0.76 & \best{0.86} & \bestErr{0.80} & 0.79 &  0.79 & 0.70 \\
        \it PSNR          & 0.61 & 0.68 & 0.52 & 0.68  & 0.78 & 0.76 & 0.75 & 0.65  & \best{0.78} & \best{0.86} & \best{0.81} & 0.83 &  0.79 & 0.73 \\
        \it SSIM          & \best{0.75} & 0.68 & 0.49 & 0.64  & 0.81 & 0.80 & 0.76 & 0.88  & 0.49 & 0.55 & 0.62 & 0.60 &  0.44 & 0.61 \\
        \it VI            & 0.57 & 0.69 & 0.43 & 0.60  & 0.69 & 0.82 & 0.67 & 0.87  & 0.59 & 0.76 & 0.68 & 0.67 &  0.41 & 0.62 \\
        \it LPIPS (2D)    & 0.63 & 0.62 & 0.35 & 0.56  & 0.76 & 0.62 & 0.87 & 0.92  & 0.71 & 0.83 & 0.79 & 0.76 &  0.87 & 0.76 \\
        \it LSiM (2D)     & 0.57 & 0.55 & 0.48 & 0.71  & 0.79 & 0.75 & 0.93 & \best{0.97}  & 0.69 & \best{0.86} & 0.79 & 0.81 &  \best{0.98} & 0.81 \\
        \it VolSiM (ours) & \best{0.75} & \best{0.73} & \best{0.66} & \best{0.77}  & \best{0.84} & \best{0.88} & \best{0.95} & \bestErr{0.96}  & \bestErr{0.77} & \best{0.86} & \best{0.81} & \best{0.88} &  0.95 & \best{0.85} \\
        
        \cmidrule(lr){1-15}
        \it CNN\textsubscript{trained}    & 0.60 & 0.71 & 0.63 & 0.76  & 0.81 & 0.77 & 0.92 & 0.93  & 0.75 & 0.86 & 0.78 & 0.85 &  0.95 & 0.82 \\
        \it MS\textsubscript{rand}        & 0.57 & 0.66 & 0.45 & 0.69  & 0.76 & 0.75 & 0.80 & 0.78  & 0.74 & 0.86 & 0.80 & 0.82 &  0.84 & 0.74 \\
        \it CNN\textsubscript{rand}       & 0.52 & 0.66 & 0.49 & 0.69  & 0.77 & 0.70 & 0.93 & 0.96  & 0.74 & 0.85 & 0.79 & 0.83 &  0.95 & 0.81 \\
        \it MS\textsubscript{identity}    & 0.75 & 0.71 & 0.68 & 0.73  & 0.83 & 0.85 & 0.87 & 0.96  & 0.74 & 0.87 & 0.77 & 0.87 &  0.94 & 0.82 \\
        \it MS\textsubscript{3 scales}    & 0.70 & 0.69 & 0.70 & 0.73  & 0.83 & 0.82 & 0.95 & 0.94  & 0.76 & 0.87 & 0.80 & 0.88 &  0.93 & 0.83 \\
        \it MS\textsubscript{5 scales}    & 0.78 & 0.72 & 0.78 & 0.78  & 0.81 & 0.90 & 0.94 & 0.93  & 0.75 & 0.85 & 0.77 & 0.88 &  0.93 & 0.82 \\
        
        \it MS\textsubscript{added \texttt{Iso}}   & 0.73 & 0.72 & 0.77 & 0.79  & 0.84 & 0.84 & 0.92 & 0.97  & \different{0.79} & 0.87 & 0.80 & 0.86 &  0.97 & 0.84 \\
        \it MS\textsubscript{only \texttt{Iso}}    & 0.58 & 0.62 & 0.32 & 0.63  & 0.78 & 0.65 & 0.72 & 0.92  & \different{0.82} & 0.77 & 0.86 & 0.79 &  0.65 & 0.75 \\
  
        \bottomrule
    \end{tabular}
    
    \begin{tablenotes}
        \item[a] Johns Hopkins Turbulence DB \cite{perlman2007_Data}
        \item[b] ScalarFlow \cite{eckert2019_ScalarFlow}
        \item[c] Combined test data sets
    \end{tablenotes}
    \end{threeparttable}

    \caption{Top: performance comparison of different metrics for 3D data via the SRCC, where values closer to $1$ indicate a better reconstruction of the ground truth distances (\best{bold+underlined}: best method for each data set, \bestErr{bold}: within a $0.01$ margin of the best performing). Bottom: ablation study of the proposed method (\different{gray}: advantage due to different training data).}
    \label{tab: results}
\end{table*}

\section{Results} \label{sec: results}
We compare the proposed \textit{VolSiM} metric to a variety of existing methods in the upper section of Tab.~\ref{tab: results}. All metrics were evaluated on the volumetric data from Sec.~\ref{sec: sequence creation}, which contain a wide range of test sets that differ strongly from the training data.
\textit{VolSiM} consistently reconstructs the ground truth distances from the entropy-based similarity model more reliably than other approaches on most data sets. As expected, this effect is most apparent on the validation sets since their distribution is closest to the training data. But even on the majority of test sets with a very different distribution, \textit{VolSiM} is the best performing or close to the best performing metric. Metrics without deep learning often fall short, indicating that they were initially designed for different use cases, like \textit{SSIM} \cite{wang2004_Image} for images, or variation of information \textit{VI} \cite{meila2007_Comparing} for clustering. The strictly element-wise metrics \textit{MSE} and \textit{PSNR} exhibit almost identical performance, and both work poorly on a variety of data sets. 
As the learning-based methods \textit{LPIPS} \cite{zhang2018_Unreasonable} and \textit{LSiM} \cite{kohl2020_Learning} are limited to two dimensions, their assessments in Tab.~\ref{tab: results} are obtained by averaging sliced evaluations for all three spatial axes. Both methods show improvements over the element-wise metrics, but are still clearly inferior to the performance of \textit{VolSiM}. This becomes apparent on our aggregated test sets displayed in the \texttt{All} column, where \textit{LSiM} results in a correlation value of 0.81, compared to \textit{VolSiM} with 0.85. \textit{LSiM} can only come close to \textit{VolSiM} on less challenging data sets where correlation values are close to 1 and all learned reconstructions are already highly accurate. This improvement is comparable to using \textit{LPIPS} over \textit{PSNR}, and represents a significant step forward in terms of a robust similarity assessment.

The bottom half of Tab.~\ref{tab: results} contains an ablation study of the proposed architecture \textit{MS}, and a simple \textit{CNN} model. This model is similar to an extension of the convolution layers of AlexNet \cite{krizhevsky2017_ImageNet} to 3D, and does not utilize a multiscale structure. Even though \textit{VolSiM} has more than $80\%$ fewer weights compared to \textit{CNN\textsubscript{trained}}, it can fit the training data more easily and generalizes better for most data sets in Tab.~\ref{tab: results}, indicating the strengths of the proposed multiscale architecture. The performance of untrained models \textit{CNN\textsubscript{rand}} and \textit{MS\textsubscript{rand}} confirm the findings from \citet{zhang2018_Unreasonable}, who also report a surprisingly strong performance of random networks.
We replace the non-linear transformation of $\vw_s$ from the similarity model with an identity transformation for \textit{MS\textsubscript{identity}} during training, i.e.~only the sequence order determines $\vg$. This consistently lowers the generalization of the metric across data sets, indicating that well calibrated sequences as well as the similarity model are important for the similarity assessment. Removing the last resolution scale block for \textit{MS\textsubscript{3 scales}} overly reduces the capacity of the model, while adding another block for \textit{MS\textsubscript{5 scales}} is not beneficial. In addition, we also investigate two slightly different training setups: for MS\textsubscript{added \texttt{Iso}} we integrate additional sequences created like the \texttt{Iso} data in the training, while MS\textsubscript{only \texttt{Iso}} is exclusively trained on such sequences. MS\textsubscript{added \texttt{Iso}} only slightly improves upon the baseline, and even the turbulence-specific MS\textsubscript{only \texttt{Iso}} model does not consistently improve the results on the JHTDB data sets. Both cases indicate a high level of generalization for \textit{VolSiM}, as it was not trained on any turbulence data.

\begin{figure}[bht]
    \centering
    \includegraphics[width=0.47\textwidth]{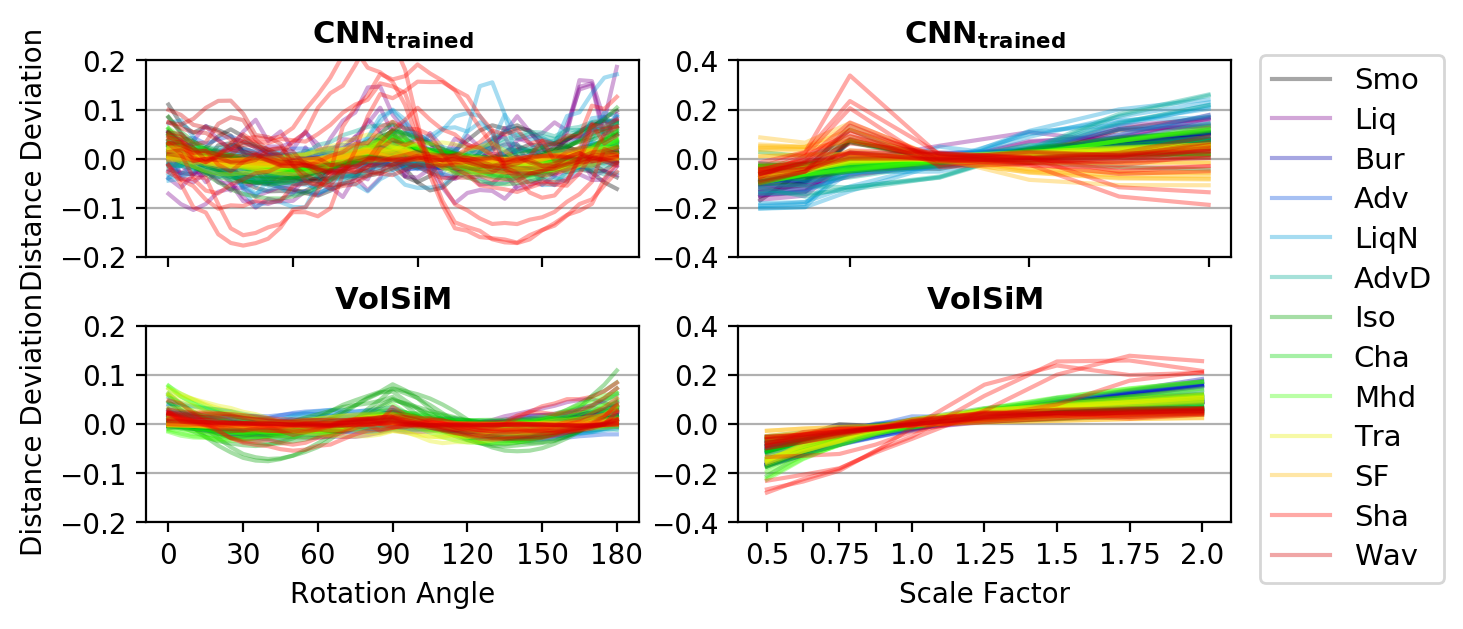}   
    \caption{Distance deviation from the mean prediction over differently rotated (left) and scaled (right) inputs for a simple CNN and the proposed multiscale model.}
    \label{fig: transformation invariance}
\end{figure}

\begin{figure*}[!t]
    \centering
    
    \includegraphics[width=0.99\textwidth]{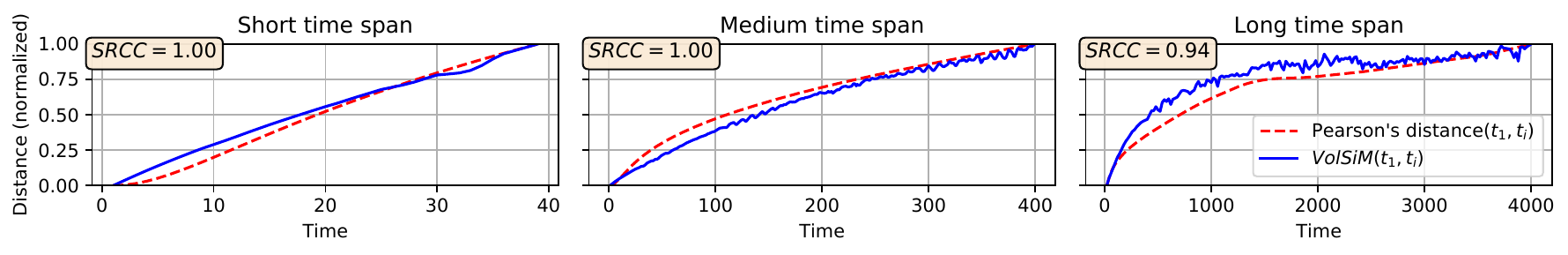}
        
    \includegraphics[width=0.99\textwidth]{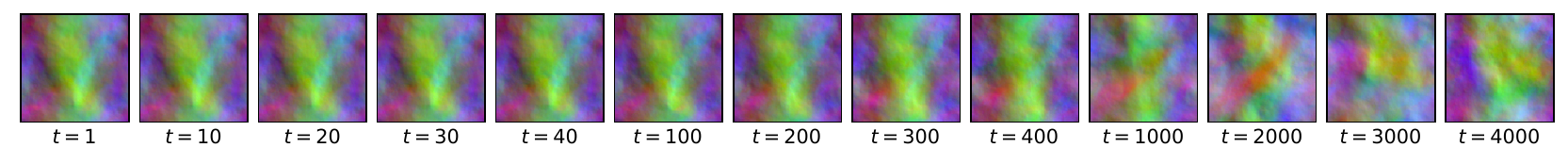}
       
    \caption{Top:~Analysis of forced isotropic turbulence across three time spans. The high SRCC values indicate strong agreement between a traditional correlation evaluation and \textit{VolSiM}. Bottom:~Examples from the sequence, visualized via a mean projection along the x-axis and color-coded channels.}
    \label{fig: isotropic case study}
\end{figure*}

\paragraph{Transformation Invariance}
Physical systems are often char\-ac\-ter\-ized by Gal\-i\-le\-an invariance \cite{mccomb1999_Dynamics}, i.e. identical laws of motion across inertial frames. Likewise, a metric should be invariant to transformations of the input, meaning a constant distance output when translating, scaling, or rotating both inputs. Element-wise metrics fulfill these properties by construction, but our Siamese network structure requires an equivariant feature representation that changes along with input transformations to achieve them. As CNN features are translation equivariant by design (apart from boundary effects and pooling), we empirically examine rotation and scale invariance for our multiscale metric and a standard Conv+ReLU model on a fixed set of 8 random data pairs from each data set. For the rotation experiment, we rotate the pairs in steps of 5\textdegree~around a random coordinate axis. The empty volume inside the original frame is filled with a value of $0$, and data outside the frame is cropped. For scaling, the data is bilinearly up- or downsampled according to the scale factor, and processed fully convolutionally.

In Fig.~\ref{fig: transformation invariance}, the resulting distance deviation from the mean of the predicted distances is plotted for rotation and scaling operations. The optimal result would be a perfectly equal distance with zero deviation across all transformations. Compared to the model \textit{CNN\textsubscript{trained}}, it can be observed that \textit{VolSiM} produces less deviations overall, and leads to significantly smoother and more consistent distance curves, across scales and rotations as shown in Fig.~\ref{fig: transformation invariance}. This is caused by the multiscale architecture, which results in a more robust internal feature representation, and thus higher stability across small transformations. Note that we observe scale equivariance rather than invariance for \textit{VolSiM}, i.e.~a mostly linear scaling of the distances according to the input size. This is most likely caused by a larger spatial size of the fully convolutional features. Making a scale equivariant model fully invariant would require a form of normalization, which is left for future work.

\paragraph{Case Study: Turbulence Analysis}
As a particularly challenging test for generalization, we further perform a case study on forced isotropic turbulence that resembles a potential real-world scenario for our metric in Fig.~\ref{fig: isotropic case study}. For this purpose, fully resolved raw DNS data over a long temporal interval from the isotropic turbulence data set from JHTDB is utilized (see bottom of Fig.~\ref{fig: isotropic case study}). The $1024^3$ domain is filtered and reduced to a size of $128^3$ via strides, meaning \textit{VolSiM} is applied in a fully convolutional manner, and has to generalize beyond the training resolution of $64^3$. Three different time spans of the simulation are investigated, where the long span also uses temporal strides. Traditionally, turbulence research makes use of established two-point correlations to study such cases \cite{pope2000_Turbulent}. Since we are interested in a comprehensive spatial analysis instead of two single points, we can make use of Pearson's distance that corresponds to an aggregated two-point correlation on the full fields to obtain a physical reference evaluation in this scenario.

Fig.~\ref{fig: isotropic case study} displays normalized distance values between the first simulation time step $t_1$ and each following step $t_i$. Even though there are smaller fluctuations, the proposed \textit{VolSiM} metric (blue) behaves very similar to the physical reference of aggregated two-point correlations (red dashed) across all time spans. This is further emphasized by the high SRCC values between both sets of trajectories, even for the challenging long time span. Our metric faithfully recovers the correlation-based reference, despite not having seen any turbulence data at training time. Overall, this experiment shows that the similarity model integrates physical concepts into the comparisons of \textit{VolSiM}, and indicates the generalization capabilities of the multiscale metric to new cases.

\section{Conclusion} \label{sec: conclusion}
We presented the multiscale CNN architecture \textit{VolSiM}, and demonstrated its capabilities as a similarity metric for volumetric simulation data. A similarity model based on the behavior of entropy in physical systems was proposed and utilized to learn a robust, physical similarity assessment. Different methods to compute correlations inside a loss function were analyzed, and the invariance to scale and rotation transformations investigated. Furthermore, we showed clear improvements upon elementwise metrics as well as existing learned approaches like \textit{LPIPS} and \textit{LSiM} in terms of an accurate similarity assessment across our data sets.

The proposed metric potentially has an impact on a broad range of disciplines where volumetric simulation data arises. An interesting area for future work is designing a metric specifically for turbulence simulations, first steps towards which were taken with our case study. Additionally, investigating learning-based methods with features that are by construction equivariant to rotation and scaling may lead to further improvements in the future.

\section*{Ethical Statement}
Since we target the fundamental problem of the similarity assessment of numerical simulations, we do not see any direct negative ethical implications of our work. However, there could be indirect negative effects since this work can act as a tool for more accurate and/or robust numerical simulations in the future, for which a military relevance exists. A further indirect issue could be explainability, e.g.~when simulations in an engineering process yield unexpected inaccuracies.

\section*{Acknowledgments}
This work was supported by the ERC Consolidator Grant \textit{SpaTe} (CoG-2019-863850).

{
\hbadness=99999 

\bibliography{bibliography}
}

\clearpage


\appendix
\section*{APPENDIX}

In the following, additional details for the proposed \textit{VolSiM} metric are provided: App.~\ref{app: implementation details} contains implementation details regarding the training and the metric model setup, and App.~\ref{app: data set details} features generation details and visualizations for all our data sets, as well as an analysis of the data distribution resulting from the data generation. The training stability for the loss experiment is investigated in App.~\ref{app: loss experiment stability}, and details for the experimental setup of the ablation study models are given in App.~\ref{app: ablation study details}. Finally, App.~\ref{app: additional ablations} contains additional ablation studies, and details regarding the turbulence case study can be found in App.~\ref{app: case study details}.

\section{Implementation Details} \label{app: implementation details}
The training and evaluation process of the metric was implemented in PyTorch \cite{paszke2019_PyTorch}, while the data was simulated and collected with specialized solvers and data interfaces as described in App~\ref{app: data set details}. The data acquisition, training, and metric evaluation was performed on a server with an Intel i7-6850 (3.60Ghz) CPU and an NVIDIA GeForce GTX 1080 Ti GPU. It took about 38 hours of training to fully optimize the final \textit{VolSiM} model for the data sequences with a spatial resolution of $64^3$.

In addition to the multiscale feature extractor network, the following operations were used for the Siamese architecture of the metric: Each feature map is normalized via a mean and standard deviation normalization to a standard normal distribution. The mean and standard deviation of each feature map is computed in a pre-processing step for the initialization of the network over all data samples. Both values are fixed for training the metric afterwards. To compare both sets of feature maps in the latent space, a simple element-wise, squared difference is employed. To keep the mathematical metric properties, this also requires a square root operation before the final distance output. The spatial squared feature map differences are then aggregated along all dimensions into a scalar distance output. Here, we used a single learned weight with dropout for every feature map, to combine them to a weighted average per network layer. The activations of the average feature maps are spatially combined with a simple mean, and summed over all network layers afterwards. This process of normalizing, comparing, and aggregating the feature maps computed by the feature extractor follows previous work \cite{kohl2020_Learning, zhang2018_Unreasonable}.

The weights to adjust the influence of each feature map are initialized to $0.1$, all other weights of the multiscale feature extractor are initialized with the default PyTorch initialization. For the final loss, the MSE term was weighted with $\lambda_1=1.0$, while the correlation term was weighted with $\lambda_2=0.7$.

\section{Data Set Details} \label{app: data set details}
In the following sections, the details underlying each data set are described. Tab.~\ref{tab: simulated data details} contains a summary of simulator, simulation setup, varied parameters, noise integration, and used fields for the simulated and generated data sets. Tab.~\ref{tab: collected data details} features a summary of the collected data sets, with repository details, jitter and cutout settings, and spatial and temporal $\Delta$ values. Both tables also contain the number of sequences created for training, validation, and testing for every data source. These values only apply for the general metric setup, and changes for the ablation study models can be found in App.~\ref{app: ablation study details} below.

\subsection{Advection-Diffusion and Burgers' Equation}
In its simplest form, the transport of matter in a flow can be described by the two phenomena of advection and diffusion. Advection describes the movement of a passive quantity inside a velocity field over time, and diffusion describes the process of dissipation of this quantity due to the second law of thermodynamics.
\begin{equation}
\frac{\partial d}{\partial t} = \nu \nabla^2 d - u \cdot \nabla d
\label{eq: advection diffusion}
\end{equation}

Eq.~\ref{eq: advection diffusion} is the simplified Advection-Diffusion equation with constant diffusivity and no sources or sinks, where $u$ denotes the velocity, $d$ is a scalar passive quantity that is transported, and $\nu$ is the diffusion coefficient or kinematic viscosity.

Burgers' Equation in Eq.~\ref{eq: burgers} is similar to the Advection-Diffusion equation, but it describes how the velocity field itself changes over time with advection and diffusion. The diffusion term can also be interpreted as a viscosity, that models the resistance of the material to deformations. Furthermore, this variation can develop discontinuities (also called shock waves). Here, $u$ also denotes the velocity and $\nu$ the kinematic viscosity or diffusion coefficient.
\begin{equation}
\frac{\partial u}{\partial t} = \nu \nabla^2 u - u \cdot \nabla u
\label{eq: burgers}
\end{equation}

To solve both PDEs, the differentiable fluid framework PhiFlow \cite{holl2020_Learning} was used. The solver utilizes a Semi-Lagrangian advection scheme, and we chose periodic domain boundary conditions to allow for the usage of a Fourier space diffusion solver. We introduced additional continuous forcing to the simulations by adding a force term $f$ to the velocity after every simulation step. Thus, $f$ depends on the time steps $t$, that is normalized by division of the simulation domain size beforehand. For \texttt{Adv}, \texttt{Bur}, and \texttt{AdvD}, we initialized the fields for velocity, density, and force with multiple layered parameterized sine functions. This leads to a large range of patterns across multiple scales and frequencies when varying the sine parameters.
\begin{equation}
    \begin{aligned}
    u^x(\vp) = \operatorname{sum} \Big( \vf_1^x + \sum_{i=1}^{4} \vf_{i+1}^x \\
    * \operatorname{sin} (2^i\pi \vp + c_{i}\,\vo_{(i+1 \operatorname{mod} 2)+1}^x) \Big)\\
    \text{where}\,\, \vc = (1,\,1,\,0.4,\,0.3)
    \label{eq: velocity init}
    \end{aligned}
\end{equation}

\vspace{0.1cm}

\begin{equation}
    \begin{aligned}
    f^x(\vp, t) = \operatorname{sum} \Big( \vf_6^x * (1+\vf_6^x*20) \\
    * \sum_{i=1}^{4} \vf_{i+1}^x * \operatorname{sin} (2^i\pi \tilde{\vp} + c_{i}\,\vo_{(i \operatorname{mod} 2)+1}^x) \Big)\\
    \text{where}\,\, \tilde{\vp}  = \vp + \vf_7^x*0.5 + \vf_7^x * \operatorname{sin} (3t) \\
    \text{and}\,\, \vc = (0,\,1,\,1,\,0.7)
    \label{eq: force init}
    \end{aligned}
\end{equation}

\begin{equation}
    d(\vp) = \operatorname{sum} \Big( \sum_{i}^{\{x,y,z\}} \operatorname{sin} (\vf_d^i * 24 \pi p_i + o_d^i) \Big)
    \label{eq: density init}
\end{equation}

Eq.~\ref{eq: velocity init}, \ref{eq: force init}, and \ref{eq: density init} show the layered sine functions in $u^x(\vp)$, $f^x(\vp)$, and $d(\vp)$ for a spatial grid position $\vp \in \mathbb{R}^3$. The $\operatorname{sum}$ operation denotes a sum of all vector elements here, and all binary operations on vectors and scalars use broadcasting of the scalar value to match the dimensions. Eq.~\ref{eq: param init} shows the definition of the function parameters used above, all of which are randomly sampled based on the simulation seed for more diverse simulations.
\begin{equation}
    \footnotesize
    \begin{aligned}
    &\vf_1^x \sim \mathcal{U}(-0.2,0.2)^3\\
    &\vf_2^x \sim \mathcal{U}(-0.2,0.2)^3\\
    &\vf_3^x \sim \mathcal{U}(-0.15,0.15)^3\\
    &\vf_4^x \sim \mathcal{U}(-0.15,0.15)^3\\
    &\vf_5^x \sim \mathcal{U}(-0.1,0.1)^3\\
    &\vf_6^x \sim \mathcal{U}(0.0,0.1)^3\\
    \end{aligned}
    \qquad
    \begin{aligned}
    &\vf_7^x \sim \mathcal{U}(-0.1,0.1)^3\\
    &\vf_d^{x,y,z} \sim \{1,\tfrac{1}{2},\tfrac{1}{3},\tfrac{1}{4},\tfrac{1}{5},\tfrac{1}{6}\}^3\\
    &\nu \sim \mathcal{U}(0.0002,0.1002)\\
    &\vo_1^x \sim \mathcal{U}(0,100)^3\\
    &\vo_2^x \sim \mathcal{U}(0,100)^3\\
    &\vo_d^{x,y,z} \sim \mathcal{U}(0,100)^3\\
    \end{aligned}
    \label{eq: param init}
\end{equation}
Note that $\nu$ was multiplied by $0.1$ for \texttt{Bur}.
The remaining velocity and force components $u^y(\vp)$, $u^z(\vp)$, $f^y(\vp)$, and $f^z(\vp)$ and corresponding parameters are omitted for brevity here, since they follow the same initialization pattern as $u^x(\vp)$ and $f^x(\vp)$. Tab.~\ref{tab: simulated data details} shows the function parameters that were varied, by using the random initializations and adjusting one of them in linear steps to create a sequence. The main difference between the Advection-Diffusion training data and the test set is the method of noise integration: For \texttt{Adv} it is integrated into the simulation velocity, while for \texttt{AdvD} it is added to the density field instead. The amount of noise added to the velocity for \texttt{Bur} and \texttt{Adv} and to the density for \texttt{AdvD} was varied in isolation as well.

\subsection{Navier-Stokes Equations}
The Navier-Stokes Equations fully describe the behavior of fluids like gases and liquids, via modelling advection, viscosity, and pressure effects, as well as mass conservation. Pressure can be interpreted as the force exerted by surrounding fluid mass at a given point, and the conservation of mass means that the fluid resists compression.
\begin{gather}
\frac{\partial u}{\partial t} + (u \cdot \nabla) u = - \frac{\nabla P}{\rho} + \nu \nabla^2 u + g
\label{eq: navier stokes momentum}
\\
\nabla \cdot u = 0.
\label{eq: navier stokes mass}
\end{gather}
Eq.~\ref{eq: navier stokes momentum} describes the conservation of momentum, and Eq.~\ref{eq: navier stokes mass} describes mass conservation. Again, $u$ denotes the velocity, $P$ is the pressure, $\rho$ is the fluids density, $\nu$ is the kinematic viscosity, and $g$ denotes external forces like gravity.

\paragraph{Smoke}
To create the smoke data set \texttt{Smo}, the fluid framework MantaFlow \cite{thuerey2018_MantaFlow} that provides a grid-based Eulerian smoke solver for the Navier-Stokes Equations was used. It is based on a Semi-Lagrangian advection scheme, and on the conjugate gradient method as a pressure solver. The simulation setup consists of a cylindrical smoke source at the bottom of the domain with a fixed noise pattern initialization to create more diverse smoke plumes. Furthermore, a constant spherical force field \textit{ff} is positioned over the source. This setup allows for a variation of multiple simulation parameters, like the smoke buoyancy, the source position and different force field settings. They include position, rotation, radius and strength. In addition, the amount of added noise to the velocity can also be varied in isolation. 

\paragraph{Liquid}
Both liquid data sets, \texttt{Liq} and \texttt{LiqN}, were created with a liquid solver in MantaFlow. It utilizes the hybrid Eulerian-Lagrangian fluid implicit particle method \cite{zhu2005_Animating}, that combines the advantages of particle and grid-based liquid simulations for reduced numerical dissipation. The simulation setup consists of two liquid cuboids of different shapes, similar to the common breaking dam setup. After 25 simulation time steps a liquid drop is added near the top of the simulation domain, and it falls down on the water surface that is still moving. Here, the external gravity force as well as the drops position and radius are varied to create similarity sequences. As for the smoke data, a modification of the amount of noise added to the velocity was also employed as a varied parameter. The main difference between the liquid training data and the test set is the method of noise integration: For \texttt{Liq} it is integrated into the simulation velocity, while for \texttt{LiqN} it is overlayed on the simulation background.

\subsection{Generated Data}
To create the shape data set \texttt{Sha} and the wave data set \texttt{Wav}, a random number of straight paths are created by randomly generating a start and end point inside the domain. It is ensured that both are not too close to the boundaries and that the path has a sufficient length. The intermediary positions for the sequence are a result of linearly interpolating on these paths. The positions on the path determine the center for the generated objects that are added to an occupancy marker grid. For both data sets, overlapping shapes and waves are combined additively, and variations with and without overlayed noise to the marker grid were created.

\paragraph{Shapes}
For \texttt{Sha}, random shapes (box or sphere) are added to the positions, where the shape's size is a random fraction of the path length, with a minimum and maximum constraint. The created shapes are then applied to the marker grid either with or without smoothed borders.

\paragraph{Waves}
For \texttt{Wav}, randomized volumetric damped cosine waves are added around the positions instead. The marker grid value $m$ at point $\vp$ for a single wave around a center $\vc$ is defined as
\begin{equation*}
    m(\vp) = cos( w * \tilde{p} ) * e^{ - (3.7\tilde{p} / r) } \; \text{where }\, \tilde{p} = \left\| \vp - \vc \right\|_2.
    \label{eq: wave init}
\end{equation*}
Here, $r$ is the radius given by the randomized size that is computed as for \texttt{Sha}, and $w \sim \mathcal{U}(0.1,0.3)$ is a randomized waviness value, that determines the frequency of the damped wave.

\subsection{Collected Data}
The collected data sets \texttt{Iso}, \texttt{Cha}, \texttt{Mhd}, and \texttt{Tra} are based on different subsets from the Johns Hopkins Turbulence Database JHTDB \cite{perlman2007_Data}, that contain different types of data from direct numerical simulations (DNS). In these simulations, all spatial scales of turbulence up to the dissipative regime are resolved. The data set \texttt{SF} is based on the ScalarFlow data \cite{eckert2019_ScalarFlow}, that contains dense 3D velocity fields of real buoyant smoke plumes, created via multi-view reconstruction technique.

\paragraph{JHTDB}
The JHTDB subsets typically contain a single simulation with a very high spatial and temporal resolution and a variety of fields. We focus on the velocity fields, since turbulent flow data is especially complex and potentially benefits most from a better similarity assessment. We can mainly rely on using temporal sequences, and only need to add spatial jitters in some cases to increase the difficulty. As turbulence generally features structures of interest across all length scales, we create sequences of different spatial scales for each subset. To achieve this, we randomly pick a cutout scale factor $s$. If $s = 1$, we directly use the native spatial discretization provided by the database. For $s > 1$ we stride the spatial query points of the normal cubical cutout by $s$ after filtering the data. For $s < 1$ the size of the cubical cutout is reduced by a factor of $s$ in each dimension, and the cutout is interpolated to the full spatial size of $128^3$ afterwards. Among other details, Tab.~\ref{tab: collected data details} shows the cutout scale factors, as well as the corresponding random weights.

\paragraph{ScalarFlow}
Since 100 reconstructions of different smoke plumes are provided in ScalarFlow, there is no need to add additional randomization to create multiple test sequences. Instead, we directly use each reconstruction sequence to create one similarity sequence in equal temporal steps. The only necessary pre-processing step is cutting off the bottom part of domain that contains the smoke inflow, since it is frequently not fully reconstructed. Afterwards, the data is interpolated to the full spatial size of $128^3$ to match the other data sets.

\subsection{Additional Example Sequences}
\begin{figure*}[hp]
    \centering
    \begin{subfigure}{\texttt{Adv}: Advection-Diffusion ($2\times$density)}
        \includegraphics[width=0.97\textwidth]{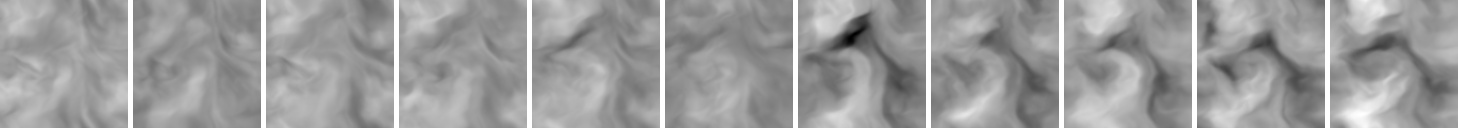}
        \includegraphics[width=0.97\textwidth]{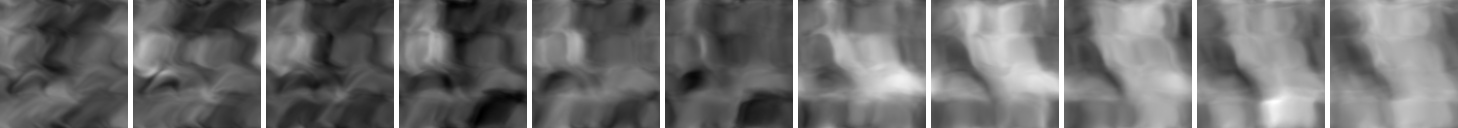}
    \end{subfigure}\\
    \vspace{0.2cm}
    
    \begin{subfigure}{\texttt{Bur}: Burgers' Equation ($2\times$velocity)}
        \includegraphics[width=0.97\textwidth]{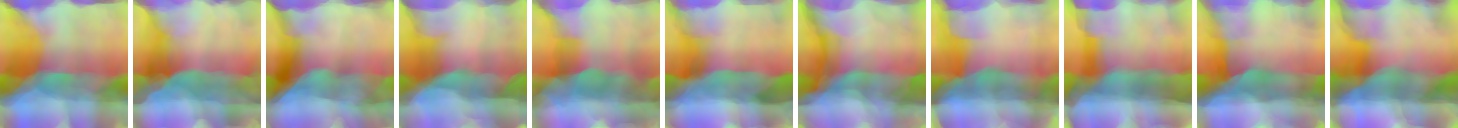}
        \includegraphics[width=0.97\textwidth]{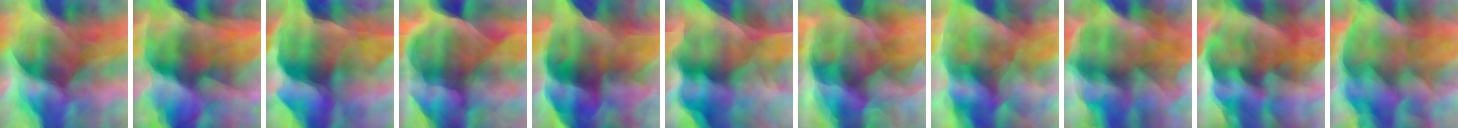}
    \end{subfigure}\\
    \vspace{0.2cm}
    
    \begin{subfigure}{\texttt{Smo}: Smoke (velocity, density, and pressure)}
        \includegraphics[width=0.97\textwidth]{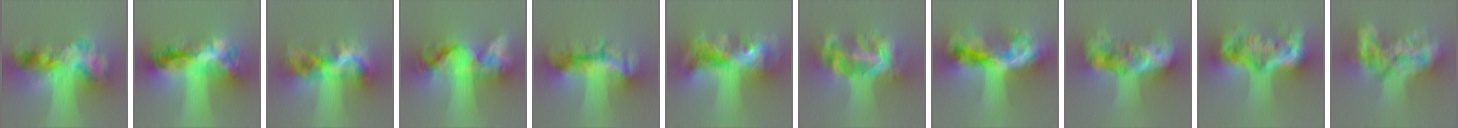}
        \includegraphics[width=0.97\textwidth]{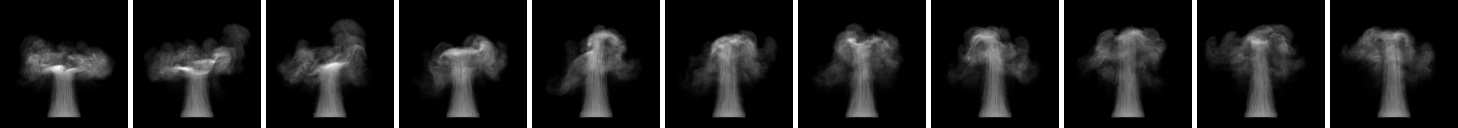}
        \includegraphics[width=0.97\textwidth]{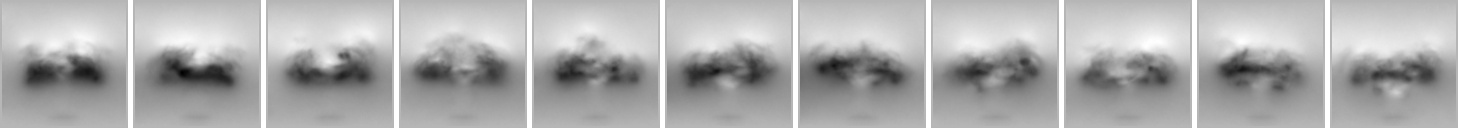}
    \end{subfigure}\\
    \vspace{0.2cm}
    
    \begin{subfigure}{\texttt{Liq}: Liquid (velocity, leveset, and flags)}
        \includegraphics[width=0.97\textwidth]{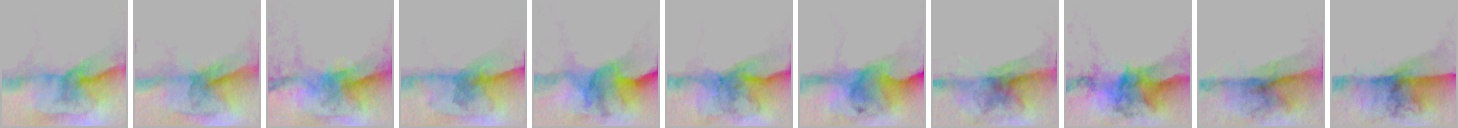}
        \includegraphics[width=0.97\textwidth]{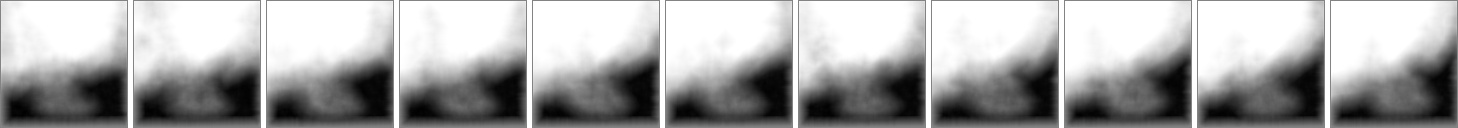}
        \includegraphics[width=0.97\textwidth]{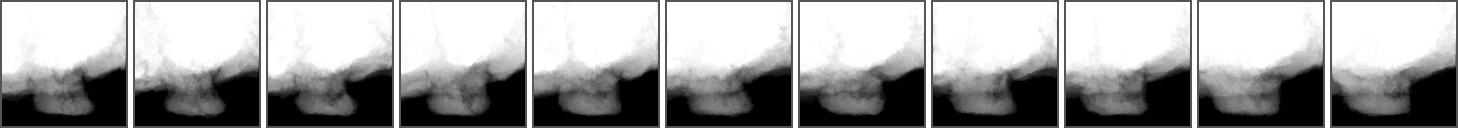}
    \end{subfigure}\\

    \caption{Example sequences of simulated training data, where each row features a full sequence from a different random seed.}
    \label{fig: training data}
\end{figure*}

\begin{figure*}[hp]
    \centering
    \begin{subfigure}{\texttt{AdvD}: Advection-Diffusion with density noise ($2\times$density)}
        \includegraphics[width=0.97\textwidth]{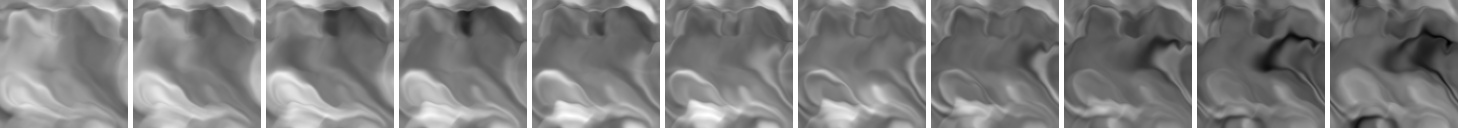}
        \includegraphics[width=0.97\textwidth]{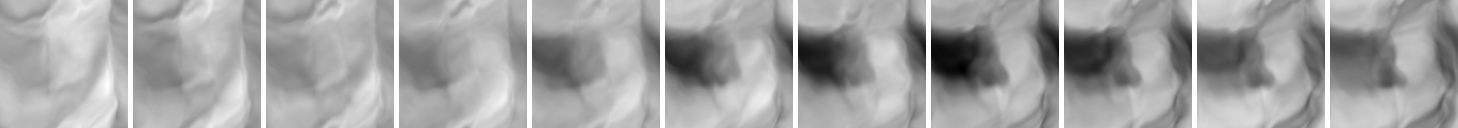}
    \end{subfigure}\\
    \vspace{0.2cm}
    
    \begin{subfigure}{\texttt{LiqN}: Liquid with background noise ($2\times$velocity)}
        \includegraphics[width=0.97\textwidth]{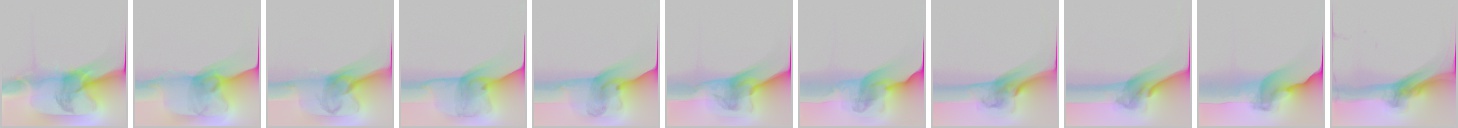}
        \includegraphics[width=0.97\textwidth]{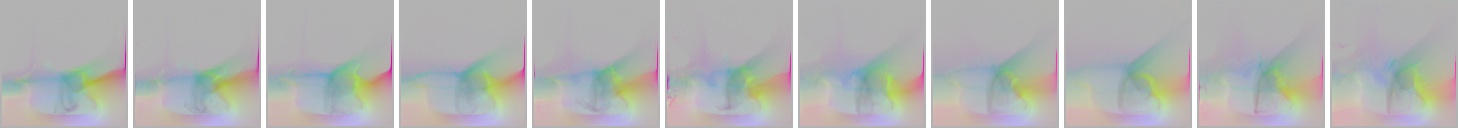}
    \end{subfigure}\\
    \vspace{0.2cm}
    
    \begin{subfigure}{\texttt{SF}: ScalarFlow ($2\times$velocity)}
        \includegraphics[width=0.97\textwidth]{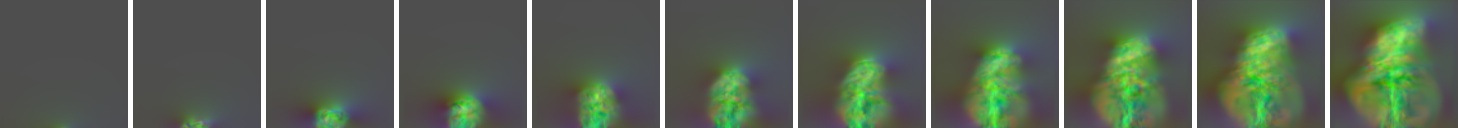}
        \includegraphics[width=0.97\textwidth]{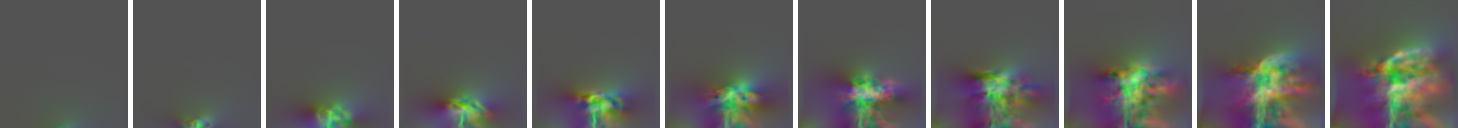}
    \end{subfigure}\\
    \vspace{0.2cm}
    
    \begin{subfigure}{\texttt{Sha}: Shapes ($2\times$marker, without and with noise)}
        \includegraphics[width=0.97\textwidth]{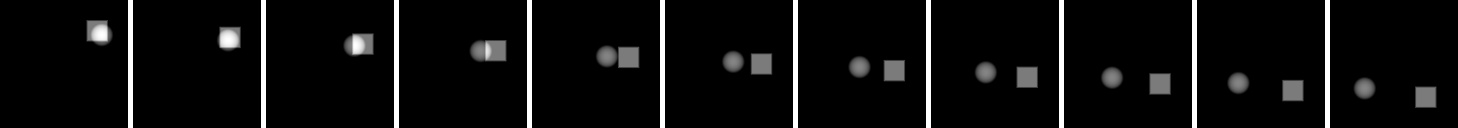}
        \includegraphics[width=0.97\textwidth]{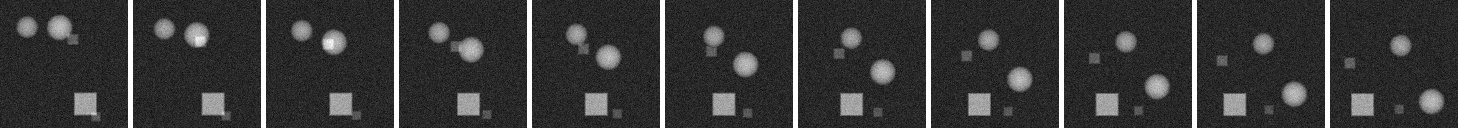}
    \end{subfigure}\\
    \vspace{0.2cm}
    
    \begin{subfigure}{\texttt{Wav}: Waves ($2\times$marker, without and with noise)}
        \includegraphics[width=0.97\textwidth]{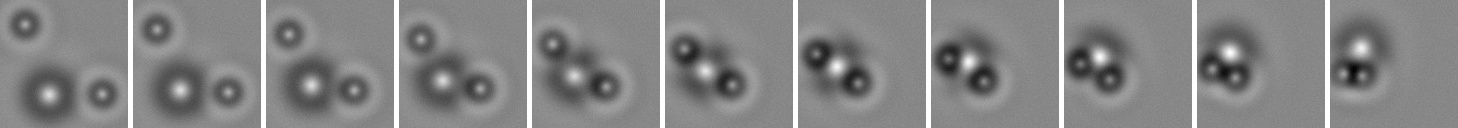}
        \includegraphics[width=0.97\textwidth]{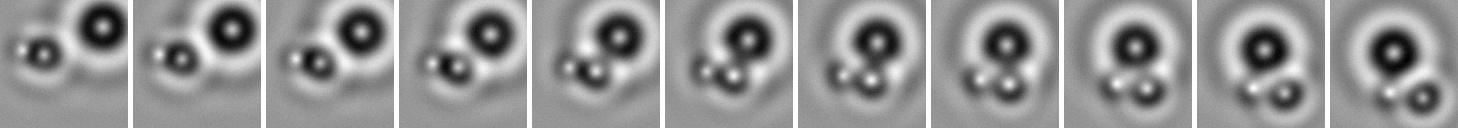}
    \end{subfigure}\\

    \caption{Example sequences of simulated (top two data sets), collected (middle data set), and generated (bottom two data sets) test data. Each row contains a full sequence from a different random seed. It is difficult to visually observe the background noise in \texttt{LiqN} due the projection along the z-axis to 2D, and due to image compression.}
    \label{fig: test data other}
\end{figure*}

\begin{figure*}[hp]
    \centering
    \begin{subfigure}{\texttt{Iso}: Isotropic turbulence ($3\times$velocity)}
        \includegraphics[width=0.94\textwidth]{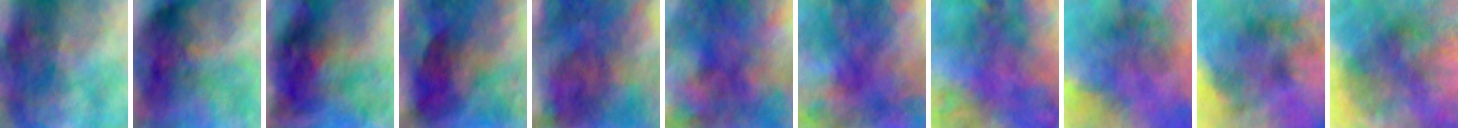}
        \includegraphics[width=0.94\textwidth]{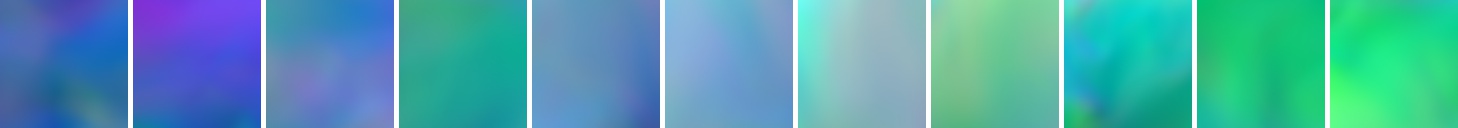}
        \includegraphics[width=0.94\textwidth]{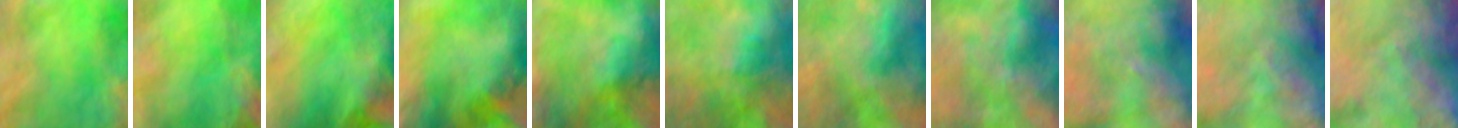}
    \end{subfigure}\\
    \vspace{0.2cm}
    
    \begin{subfigure}{\texttt{Cha}: Channel flow ($3\times$velocity)}
        \includegraphics[width=0.94\textwidth]{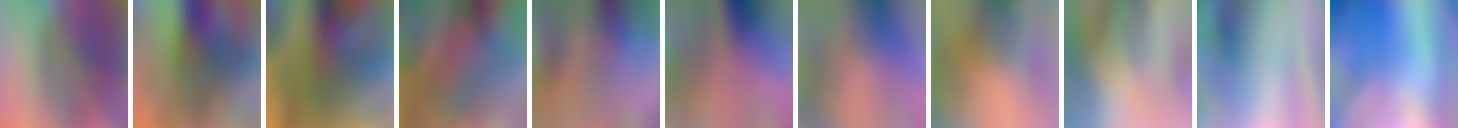}
        \includegraphics[width=0.94\textwidth]{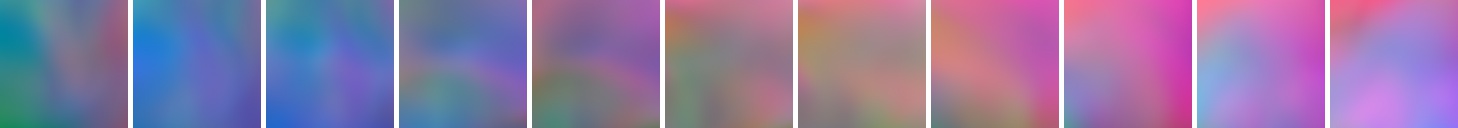}
        \includegraphics[width=0.94\textwidth]{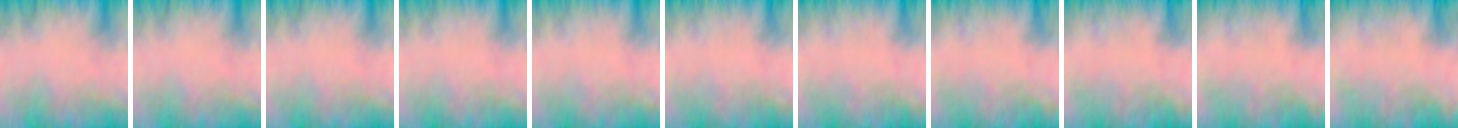}
    \end{subfigure}\\
    \vspace{0.2cm}
    
    \begin{subfigure}{\texttt{Mhd}: Magneto-hydrodynamic turbulence ($3\times$velocity)}
        \includegraphics[width=0.94\textwidth]{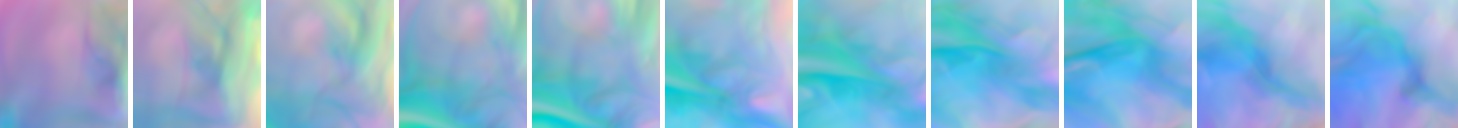}
        \includegraphics[width=0.94\textwidth]{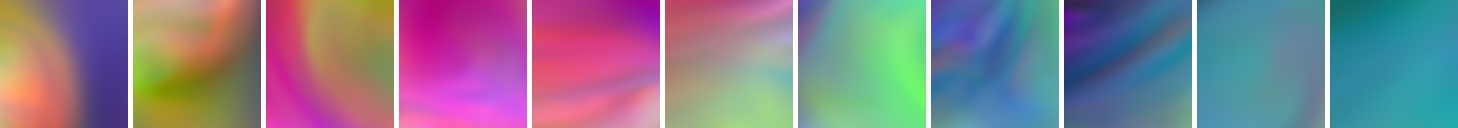}
        \includegraphics[width=0.94\textwidth]{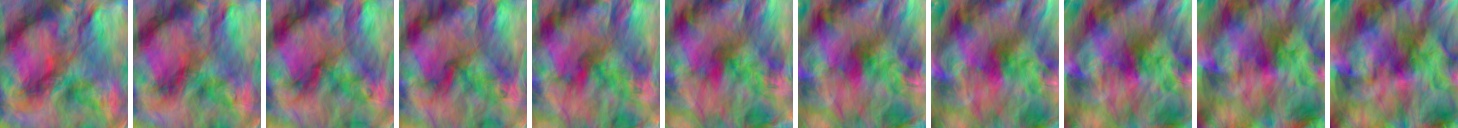}
    \end{subfigure}\\
    \vspace{0.2cm}
        
    \begin{subfigure}{\texttt{Tra}: Transitional boundary layer ($3\times$velocity)}
        \includegraphics[width=0.94\textwidth]{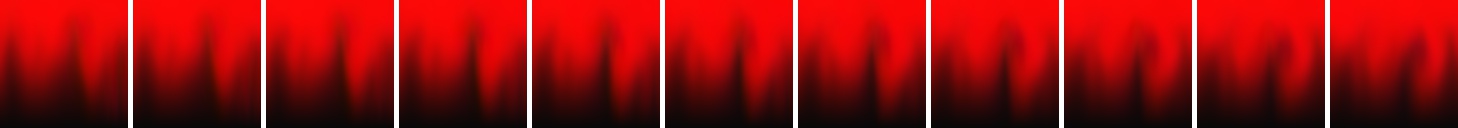}
        \includegraphics[width=0.94\textwidth]{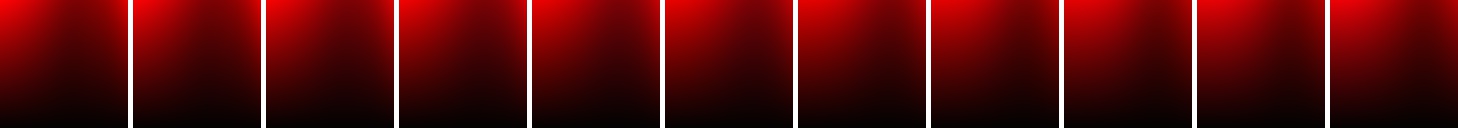}
        \includegraphics[width=0.94\textwidth]{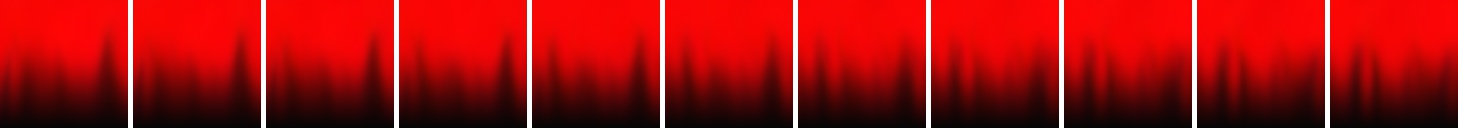}
    \end{subfigure}\\
    
    \caption{Example sequences of collected test data from JHTDB, where each row shows a full sequence from a different random seed. Notice the smaller cutout scale factor $s$ for the middle example in each case. The predominant x-component in \texttt{Cha} is separately normalized for a more clear visualization.}
    \label{fig: test data jhtdb}
\end{figure*}

Fig.~\ref{fig: training data},~\ref{fig: test data other}, and~\ref{fig: test data jhtdb} show multiple full example sequences from all our data sets. In every sequence, the leftmost image is the baseline field. Moving further to the right, the change of one initial parameter increases for simulated data sets, and the spatio-temporal position offset increases for generated and collected data. To plot the sequences, the 3D data is projected along the z-axis to 2D via a simple mean operation. This means, noise that was added to the data or the simulation is typically significantly less obvious due to statistical averaging in the projection. Velocity data is directly mapped to RGB color channels, and scalar data is shown via different shades of gray. Unless note otherwise, the data is jointly normalized to $[0,1]$ for all channels at the same time, via the overall minimum and maximum of the data field.

\subsection{Data Distribution}
There are many independent factors that influence the actual difficulty of the created data sequences. For example, our iterative data generation methods only calibrate $\Delta$ with the help of proxy functions. Furthermore, the clear correlation thresholds are only used for a small set of sequences, instead of sampling every sequence based on them.
\begin{figure}[ht]
    \centering
    \includegraphics[width=0.42\textwidth]{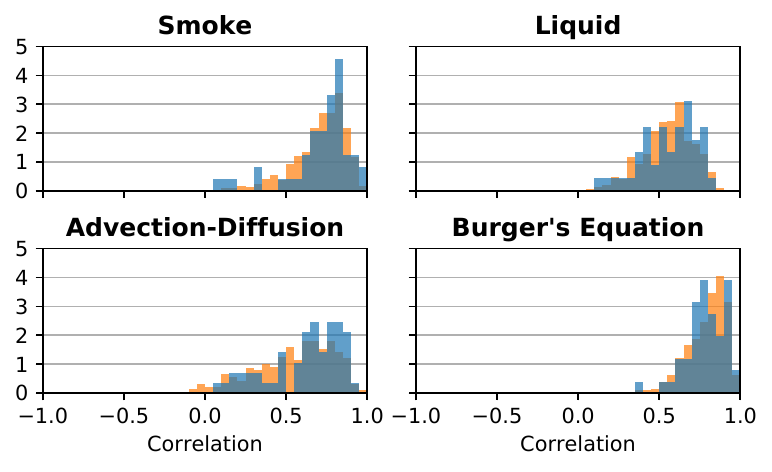}
    \caption{Normalized MSE correlation histograms of training (orange) and validation data (blue) with a bin size of $0.05$, all of which roughly follow a truncated normal distribution.}
    \label{fig: correlation histogram}
\end{figure}

As a result, the computed PCC values from the MSE (see Fig.~\ref{fig: data iteration schemes}) on the full data sets exhibit a natural, smooth distribution with controllable difficulty, instead of introducing an artificial distribution with hard cutoffs. According to the central limit theorem and taking into account that correlations have an upper bound of $1$, we expect the PCC values to follow a truncated normal distribution. Intuitively, this corresponds to a distribution of trajectories with different curvature in the similarity model in Fig.~\ref{fig: entropy behavior}. In fact, we empirically determined that only training data distributions in the PCC space which reasonably closely follow a truncated normal distribution with a sufficiently positive peak, result in a successful training of our model. We assume, that cutoffs or significantly different distributions indicate unwanted biases in the data, which prevented effective learning in our experiments. Fig.~\ref{fig: correlation histogram} shows normalized correlation histograms of our training and validation data sets, all of which roughly follow this desired truncated normal distribution.

\section{Training Stability for Loss Experiment} \label{app: loss experiment stability}
\begin{figure*}[ht]
    \centering
    \includegraphics[width=0.97\textwidth]{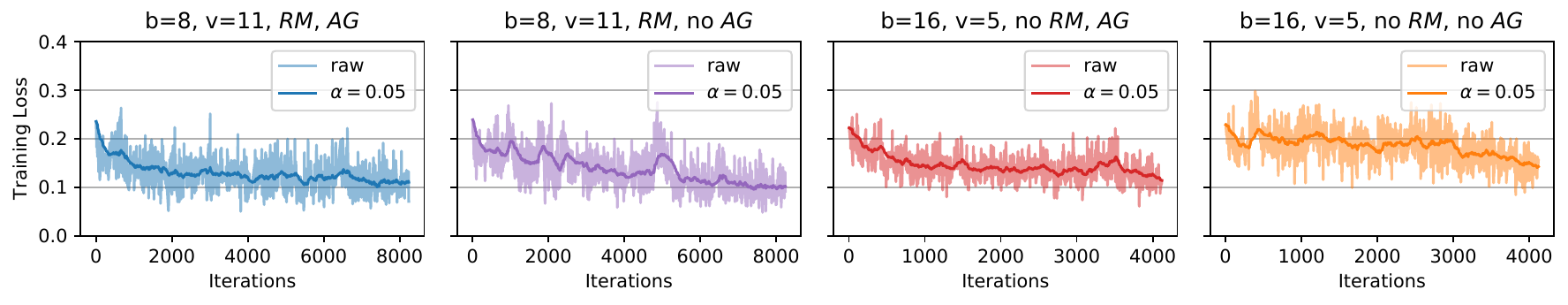}
    \caption{Loss curves from the loss function analysis in Fig.~\ref{fig: loss comparison} for different models. Shown are the raw losses per batch over all training iterations (lighter line), and an exponential weighted moving average with $\alpha=0.05$ (darker line) for better visual clarity.}
    \label{fig: loss comparison train}
\end{figure*}

To further analyze the resulting stability of different variants of the loss investigated in Sec.~\ref{sec: distance function}, Fig.~\ref{fig: loss comparison train} shows training trajectories of multiple models. Displayed are the direct training loss over the training iterations, i.e.~one loss value per batch, and a smoothed version via an exponential weighted moving average computed with $\alpha=0.05$, corresponding to a window size of 40. The models are color-coded according to Fig.~\ref{fig: loss comparison}, and were trained with different batch sizes $b$, slicing values $v$, usage of the running sample mean $RM$, and usage of the correlation aggregation $AG$. Note that the models were generally trained with the same training seed, meaning the order of samples is very similar across all runs (only with potential deviations due to race conditions and GPU processing).

In Fig.~\ref{fig: loss comparison train}, it can be observed that models trained with lower batch sizes achieve lower training losses, leading to the better generalization on test and validation sets in Fig.~\ref{fig: loss comparison}. Using no $AG$ leads to a less stable training procedure and a higher loss during most of training, especially for lower batch sizes. Occasionally, even large loss spikes occur that eradicate most training progress.

\section{Ablation Study Details} \label{app: ablation study details}
In the following, details for the ablation study models in Sec.~\ref{sec: results} are provided. The proposed \textit{VolSiM} metric uses the multiscale architecture \textit{MS} described in Sec.~\ref{sec: distance function} as a feature extractor. For all models, the general training setup mentioned in App.~\ref{app: implementation details} stays identical, apart from: 1) removing the contribution of the entropy-based similarity model, 2) changes to the feature extractor, 3) a different feature extractor, 4) the training amount, or 5) the training data.

\paragraph{No Similarity Model}
The MS\textsubscript{identity} model does not directly make use of the similarity model based on entropy. For that, the logarithmic transformation of $\vw_s = (j-i)/n$ according to the similarity model is replaced with an identity transformation during training. This corresponds to linear ground truth distances according to the order of each sequence determined by our data generation. Note that this simplified network version is slightly more efficient as the computation of the parameter $c$ can be omitted. However, pre-computing these values for all sequences is computationally very light compared to training the full metric network.

\paragraph{Changes to \textit{MS} Architecture}
For \textit{MS\textsubscript{3 scales}}, the last scale block is removed, meaning the architecture from Fig.~\ref{fig: multiscale network} ends with the $1/4$ resolution level, while the $1/8$ resolution level is omitted. 
For the \textit{MS\textsubscript{5 scales}} model, one additional scale block is added. In Fig.~\ref{fig: multiscale network}, this corresponds to a $1/16$ resolution level with a $16\times16\times16$ AvgPool and a scale block with 64 channels.

\paragraph{Simple CNN Feature Extractor}
The \textit{CNN\textsubscript{trained}} model employs an entirely different feature extractor network, similar to the classical AlexNet architecture \cite{krizhevsky2017_ImageNet}. It consists of 5 layers, each with a 3D convolution followed by a ReLU activation. The kernel sizes (12 - 5 - 3 - 3 - 3) decrease along with the strides (4 - 1 - 1 - 1 - 1), while the number of features first increase, then decreases again (32 - 96 - 192 - 128 - 128). To create some spatial reductions, two $4\times4\times4$ MaxPools with stride 2 are included before the second and third convolution. The normalization and aggregation of the resulting feature maps is performed as described in App.~\ref{app: implementation details}, in the same way as for the proposed feature extractor network. As a result, \textit{CNN\textsubscript{trained}} can also be applied in a fully convolutional manner for the scaling invariance experiment in Fig.~\ref{fig: transformation invariance}.

\begin{table*}[ht]
    \footnotesize  
    \centering

    \begin{tabular}[b]{l c c c c || c c c c c c c c c | c}
        \toprule
        \multirow{2}{*}[-1.3mm]{} & \multicolumn{4}{c ||}{\bf Validation data sets} & \multicolumn{10}{c}{\bf Test data sets} \\
        \cmidrule(lr){2-5} \cmidrule(lr){6-15}
        \bf Metric & \texttt{Adv} & \texttt{Bur} & \texttt{Liq} & \texttt{Smo} & \texttt{AdvD} & \texttt{LiqN} & \texttt{Sha} & \texttt{Wav} & \texttt{Iso} & \texttt{Cha} & \texttt{Mhd} & \texttt{Tra} & \texttt{SF} & \texttt{All}\\
        \cmidrule(lr){1-15}
        
        \it VolSiM                      & 0.75 & 0.73 & 0.66 & 0.77  & 0.84 & 0.88 & 0.95 & 0.96  & 0.77 & 0.86 & 0.81 & 0.88 &  0.95 & 0.85 \\
        \it MS\textsubscript{no skip}   & 0.80 & 0.70 & 0.78 & 0.75  & 0.86 & 0.88 & 0.80 & 0.95  & 0.76 & 0.86 & 0.78 & 0.86 &  0.91 & 0.82 \\
        \cmidrule(lr){1-15}
        
        \it MS\textsubscript{4 scales}  & 0.74 & 0.74 & 0.86 & 0.77  & 0.82 & 0.83 & 0.93 & 0.97  & 0.77 & 0.88 & 0.82 & 0.89 &  0.95 & 0.84 \\
        \it MS\textsubscript{1 scale}   & 0.73 & 0.71 & 0.78 & 0.75  & 0.83 & 0.76 & 0.91 & 0.96  & 0.73 & 0.87 & 0.78 & 0.88 &  0.94 & 0.82 \\

        \cmidrule(lr){1-15}
        \it MS\textsubscript{with pool}  & 0.70 & 0.70 & 0.76 & 0.72  & 0.82 & 0.79 & 0.94 & 0.96  & 0.76 & 0.83 & 0.79 & 0.88 &  0.97 & 0.83 \\
        \it MS\textsubscript{no pool}    & 0.73 & 0.70 & 0.72 & 0.71  & 0.83 & 0.77 & 0.94 & 0.95  & 0.76 & 0.87 & 0.78 & 0.86 &  0.90 & 0.82 \\
        \cmidrule(lr){1-15}

        \it MSE                         & 0.61 & 0.70 & 0.51 & 0.68  & 0.77 & 0.76 & 0.75 & 0.65  & 0.76 & 0.86 & 0.80 & 0.79 &  0.79 & 0.70 \\
        \it PCC                         & 0.65 & 0.69 & 0.55 & 0.65  & 0.72 & 0.80 & 0.72 & 0.73  & 0.70 & 0.83 & 0.75 & 0.78 &  0.89 & 0.69 \\

        \bottomrule
    \end{tabular}

    \caption{Performance comparison of further ablation study models via the SRCC. Shown are a model where the resolution skip connections in the feature extractor are omitted (top block), and two models without a multiscale architecture (second and third block). The upper row in each block is the corresponding baseline. Furthermore, using Pearson's distance directly as a metric is compared to an MSE evaluation (bottom block).}
    \label{tab: additional ablations}
\end{table*}

\paragraph{Random Models}
For the random models \textit{CNN\textsubscript{rand}} and \textit{MS\textsubscript{rand}}, the corresponding feature extractor along with the aggregation weights are only initialized and not trained. However, to normalize every feature to a standard normal distribution, the training data is processed once to determine the feature mean and standard deviation before evaluating the models, as detailed in App.~\ref{app: implementation details}.

\paragraph{Different Training Data}
For the MS\textsubscript{added \texttt{Iso}} model, we created 400 additional sequences from the JHTDB according to the \texttt{Iso} column in Tab.~\ref{tab: collected data details}, that are added to the other training data. Note that they utilize different random seeds than the 60 test sequences, so the \texttt{Iso} data set becomes an additional validation set in Tab.~\ref{tab: results}.
Similarly for the MS\textsubscript{only \texttt{Iso}} model, 1000 sequences according to the \texttt{Iso} column in Tab.~\ref{tab: collected data details} were collected. They replace all other training data from the original model, meaning the \texttt{Iso} data set becomes a validation set, and \texttt{Adv}, \texttt{Bur}, \texttt{Liq}, and \texttt{Smo} become further test sets in Tab.~\ref{tab: results}. For consistency, the \texttt{All} column in Tab.~\ref{tab: results} still reports the combined values of the original test sets for both cases.

\section{Additional Ablations} \label{app: additional ablations}
To further investigate our method, we perform three additional ablations in the following. The first ablation focuses on the benefits of the resolution skip connections in our multiscale feature extractor. The other two replace the multiscale aspect in different ways, to demonstrate the robustness and memory efficiency of the proposed architecture. Additionally, directly using Pearson's distance on our data is compared to an analysis using the MSE. Tab.~\ref{tab: additional ablations} shows the resulting SRCC values for these models and the corresponding baselines on our data sets, which are computed as for Tab.~\ref{tab: results}.

\paragraph{No Skip Connections}
We compare the \textit{VolSiM} model as proposed in the main paper with MS\textsubscript{no skip} that does not make use of the resolution skip connections (red dashed arrows in Fig.~\ref{fig: multiscale network}). This means each resolution is isolated, and sharing information across features of different levels is more difficult, even though all features from each layer are still used for the final distance computation. The results in the first block in Tab.~\ref{tab: additional ablations} show that removing skip connections simplifies learning to some degree as the performance clearly increases on two validation sets. However, the generalization to data that is very different compared to the training distribution becomes more difficult. This is especially obvious for the data sets \texttt{Sha} and \texttt{SF}, both of which feature relatively large visual changes. Interestingly, the effects on the \texttt{Wav} data set with similar characteristics are only minor.

\paragraph{Singlescale Model}
To investigate the multiscale aspect of our architecture, we removed all scale blocks at lower resolutions, such that only the component at the native input resolution remains (see Fig.~\ref{fig: multiscale network}). To compensate for the lower number of network parameters, we scale the number channels in each layer by a factor of $11$, leading to the rather shallow but wide network \textit{MS\textsubscript{1 scale}} with around $360k$ weights. The original network structure \textit{MS\textsubscript{4 scales}} with the four scale blocks reconstructs the ground truth more accurately for almost all data sets as displayed in the middle block of Tab.~\ref{tab: additional ablations}. Furthermore, the proposed structure requires about five times less memory during training.

\paragraph{No Pooling Layers}
A different approach to analyze the multiscale aspect, is to eliminate all AvgPool layers and set all convolution strides to $1$ (also see Fig.~\ref{fig: multiscale network}). Here, no further adjustments to the network are required as neither pooling layers nor strides alter the number of network weights. The resulting \textit{MS\textsubscript{no pool}} model is only slightly worse compared to the baseline, as indicated by bottom block in Tab.~\ref{tab: additional ablations}. However, the \textit{MS\textsubscript{with pool}} model needs about ten times less memory compared to the adjust variant during training due to the significantly smaller feature sizes.

Note that for both ablations on the multiscale structure, the training and test data resolution was reduced to $32^3$. Furthermore, the number of channels in each layer for the \textit{MS\textsubscript{with pool}} and \textit{MS\textsubscript{no pool}} models was reduced by a factor of $0.5$. Both choices are motivated by memory limitations and are the reason for the slightly different results across the three baseline models in Tab.~\ref{tab: additional ablations}.

\paragraph{Further Element-wise Metrics}
The last block in Tab.~\ref{tab: additional ablations} analyzes the performance of Pearson's distance on our data sets. Here, $\vd$ is computed via $d_i = 1 - \left|\operatorname{PCC}(s_0,s_i)\right|$ and compared it to $\vg$ via the SRCC in the same way as for the other metrics. The resulting performance is overall quite similar to an MSE evaluation, which is expected as Pearson's distance essentially still is an element-wise comparison, even though it does take some general data statistics into account.

\section{Turbulence Case Study Details} \label{app: case study details}
For the case study in Sec.~\ref{sec: results}, the velocity field from the \textit{isotropic1024coarse} data set from the JHTDB \cite{perlman2007_Data} is utilized. Instead of calibrating sequences according to data generation approach and the similarity model, the data is directly converted to three sequences of different time spans without any randomization in this case. A spatial stride of 8 is employed is employed for all cases, and to reduce memory consumption an additional temporal stride of 20 is used for the long time span only. The resulting sequences exhibit a spatial resolution of $128^3$, with 40 frames for the short span, 400 frames for medium span, and 200 frames for the long span. Before further processing, each frame is individually normalized to $[-1,1]$. To create the results in Fig.~\ref{fig: isotropic case study}, the first simulation frame $t_1$ is individually compared to all following simulation frames $t_i$ via Pearson's distance (red dashed trajectory) and \textit{VolSiM} (blue solid trajectory). Note that \textit{VolSiM} is applied fully convolutionally as it was trained on $64^3$ data, and has to generalize to $128^3$ here. For both cases, the resulting distances are normalized to $[0,1]$ to visually compare them more easily. The examples from the sequences are visualized as described in App.~\ref{app: data set details}, while projecting along the x-axis.

\begin{table*}[hp]
    \footnotesize
    \centering

    \begin{threeparttable}
    \begin{tabular}[b]{>{\raggedright\arraybackslash}m{1.73cm} | >{\centering\arraybackslash}m{1.43cm} >{\centering\arraybackslash}m{1.43cm} >{\centering\arraybackslash}m{1.43cm} >{\centering\arraybackslash}m{1.43cm} >{\centering\arraybackslash}m{1.43cm} >{\centering\arraybackslash}m{1.43cm} >{\centering\arraybackslash}m{1.43cm} >{\centering\arraybackslash}m{1.43cm}}
        \toprule
        & \texttt{Adv} & \texttt{Bur} & \texttt{Liq} & \texttt{Smo} & \texttt{AdvD} & \texttt{LiqN} & \texttt{Sha} & \texttt{Wav}\\
        \cmidrule(lr){1-9}
        
        Sequences\hspace{0.5cm}train--val--test  & 398--57--0 & 408--51--0 & 405--45--0 & 432--48--0  & 0--0--57 & 0--0--30 & 0--0--60 & 0--0--60 \\
        \cmidrule(lr){1-9}

        Equation  & Eq.~\ref{eq: advection diffusion} & Eq.~\ref{eq: burgers} & Eq.~\ref{eq: navier stokes momentum}, \ref{eq: navier stokes mass} & Eq.~\ref{eq: navier stokes momentum}, \ref{eq: navier stokes mass}   & Eq.~\ref{eq: advection diffusion} & Eq.~\ref{eq: navier stokes momentum}, \ref{eq: navier stokes mass}  & --- & --- \\
        \cmidrule(lr){1-9}

        Simulator & PhiFlow\tnote{d} & PhiFlow\tnote{d} & MantaFlow\tnote{e} & MantaFlow\tnote{e}  & PhiFlow\tnote{d} & MantaFlow\tnote{e} & MantaFlow\tnote{e} & MantaFlow\tnote{e} \\
        \cmidrule(lr){1-9}

        Simulation setup  & layered sines & layered sines  & breaking dam + drop & rising plume with force field & layered sines & breaking dam + drop & random shapes & random damped waves\\
        \cmidrule(lr){1-9}

        Time steps  & $120$ & $120$ & $80$ & $120$  & $120$ & $80$ & --- & ---\\
        \cmidrule(lr){1-9}

        Varied aspects  &
            $f_1,f_2,$ $f_3,f_4,$ $f_5,f_7,$ $o_1,o_2,$ $o_d,$ $noise$ & 
            $f_1,f_2,$ $f_3,f_4,$ $f_5,f_7,$ $o_1,o_2,$ $noise$ &
            $drop_x$ $drop_y $ $drop_z$ $drop_{rad}$ $grav_x$ $grav_y$ $grav_z$ $noise$ & 
            $buoy_x$ $buoy_y$ $ff_{rot\,x}$ $ff_{rot\,z}$ $ff_{str\,x}$ $ff_{str\,z}$ $ff_{pos\,x}$ $ff_{pos\,y}$ $ff_{rad}$ $source_x$ $source_y$ $noise$  & 
            $f_1,f_2,$ $f_3,f_4,$ $f_5,f_7,$ $o_1,o_2,$ $o_d,$ $noise$ & 
            $drop_x$ $drop_y $ $drop_z$ $drop_{rad}$ $grav_x$ $grav_y$ $grav_z$ $noise$ & 
            shape position & wave position\\
        \cmidrule(lr){1-9}

        Noise integration  & added to velocity & added to velocity & added to velocity & added to velocity  & added to density & overlay on non-liquid & overlay on marker & overlay on marker \\
        \cmidrule(lr){1-9}

        Used fields & density & velocity & velocity flags levelset & density pressure velocity  & density & velocity & marker & marker \\

        \bottomrule
    \end{tabular}
    
    \begin{tablenotes}
        \item[d] PhiFlow (\url{https://github.com/tum-pbs/PhiFlow}) from \citet{holl2020_Learning}
        \item[e] MantaFlow (\url{http://mantaflow.com/}) from \citet{thuerey2018_MantaFlow}
    \end{tablenotes}
    \caption{Data set detail summary for the simulated and generated data sets. }
    \label{tab: simulated data details}
    \end{threeparttable}
    \vspace{0.4cm}
\end{table*}

\begin{table*}[hp]
    \centering
    \footnotesize

    \begin{threeparttable}
    \begin{tabular}[b]{>{\raggedright\arraybackslash}m{2.7cm} | >{\centering\arraybackslash}m{2.2cm} >{\centering\arraybackslash}m{2.2cm} >{\centering\arraybackslash}m{2.2cm} >{\centering\arraybackslash}m{2.2cm} >{\centering\arraybackslash}m{2.2cm}}
        \toprule
        & \texttt{Iso} & \texttt{Cha} & \texttt{Mhd} & \texttt{Tra}  & \texttt{SF}\\
        \cmidrule(lr){1-6}

        Sequences\hspace{0.5cm}train--val--test  & 0--0--60 & 0--0--60 & 0--0--60 & 0--0--60  & 0--0--100 \\
        \cmidrule(lr){1-6}

        Repository  & JHTDB -- isotropic 1024coarse \tnote{f} & JHTDB -- channel \tnote{f} & JHTDB -- mhd1024 \tnote{f} & JHTDB -- transition\_bl \tnote{f}  & ScalarFlow \tnote{g} \\
        \cmidrule(lr){1-6}

        Repository size \tnote{h} $s \times t \times x \times y \times z$  & $1 \times 5028 \times 1024 \times 1024 \times 1024$ &  $1 \times 4000 \times 2048 \times 512 \times 1536$ &  $1 \times 1024 \times 1024 \times 1024 \times 1024$ & $1 \times 4701 \times 10240 \times 1536 \times 2048$  &  $100 \times 150 \times 100 \times 178 \times \quad 100$ \tnote{i} \\
        \cmidrule(lr){1-6}

        Temporal offset $\Delta_t$  & $180$ & $37$ & $95$ & $25$  & $13$ \\
        \cmidrule(lr){1-6}

        Spatial offset $\Delta_{x,y,z}$  & $0$ & $0$ & $0$ & $0$  & $0$ \\
        \cmidrule(lr){1-6}

        Spatial jitter  & $0$ & $0$ & $25$ & $0$  & $0$ \\
        \cmidrule(lr){1-6}

        Cutout scales  & $0.25,0.5,$ $0.75,1,$ $2,3,4$ & $0.25,0.5,$ $0.75,1,$ $2,3,4$ & $0.25,0.5,$ $0.75,1,$ $2,3,4$ & $0.25,0.5,$ $0.75,1,2$  & $1$ \\
        \cmidrule(lr){1-6}

        Cutout scale random weights  & $0.14,0.14,$ $0.14,0.16,$ $0.14,0.14,$ $0.14$ & $0.14,0.14,$ $0.14,0.16,$ $0.14,0.14,$ $0.14$ & $0.14,0.14,$ $0.14,0.16,$ $0.14, 0.14,$ $0.14$ & $0.14,0.14,$ $0.14,0.30,$ $0.28$  & $1$ \\
        \cmidrule(lr){1-6}

        Used fields  & velocity & velocity & velocity & velocity  & velocity \\
        \bottomrule
    \end{tabular}
    
    \begin{tablenotes}
        \item[f] JHTDB (\url{http://turbulence.pha.jhu.edu/}) from \citet{perlman2007_Data}
        \item[g] ScalarFlow (\url{https://mediatum.ub.tum.de/1521788}) from \citet{eckert2019_ScalarFlow}
        \item[h] simulations $s$ $\times$ time steps $t$ $\times$ spatial dimensions $x,y,z$
        \item[i] cut to $100 \times 150 \times 100 \times 160 \times 100$ (removing 18 bottom values from y), since the smoke inflow is not fully reconstructed
    \end{tablenotes}
    \caption{Data set detail summary for collected data sets.}
    \label{tab: collected data details}
    \end{threeparttable}
\end{table*}

\end{document}

%% file: math_commands.tex
\usepackage{amsmath,amsfonts,bm}

\def\eqref#1{equation~\ref{#1}}

\def\1{\bm{1}}

\def\vc{{\bm{c}}}
\def\vd{{\bm{d}}}

\def\vf{{\bm{f}}}
\def\vg{{\bm{g}}}

\def\vo{{\bm{o}}}
\def\vp{{\bm{p}}}

\def\vs{{\bm{s}}}

\def\vw{{\bm{w}}}

\DeclareMathAlphabet{\mathsfit}{\encodingdefault}{\sfdefault}{m}{sl}
\SetMathAlphabet{\mathsfit}{bold}{\encodingdefault}{\sfdefault}{bx}{n}
\newcommand{\tens}[1]{\bm{\mathsfit{#1}}}

\def\tD{{\tens{D}}}

